\documentclass[letterpaper]{article}
\usepackage[preprint]{aaai2027}
\usepackage[hyphens]{url}  % DO NOT CHANGE THIS
\usepackage{graphicx} % DO NOT CHANGE THIS
\usepackage{natbib}  % DO NOT CHANGE THIS AND DO NOT ADD ANY OPTIONS TO IT
\usepackage{caption} % DO NOT CHANGE THIS AND DO NOT ADD ANY OPTIONS TO IT
\usepackage{algorithm}
\usepackage{algorithmic}

\newcommand{\dvA}[2]{#1{\small\textcolor{gray}{(#2)}}}
\AtBeginDocument{%
  \setlength{\abovedisplayskip}{6pt plus 2pt minus 4pt}%
  \setlength{\belowdisplayskip}{6pt plus 2pt minus 4pt}%
  \setlength{\abovedisplayshortskip}{0pt plus 1pt}%
  \setlength{\belowdisplayshortskip}{4pt plus 2pt minus 2pt}%
  \setlength{\textfloatsep}{4pt plus 1pt minus 2pt}%
  \setlength{\floatsep}{3pt plus 1pt minus 1pt}%
  \setlength{\intextsep}{4pt plus 1pt minus 2pt}%
  \setlength{\dbltextfloatsep}{4pt plus 2pt minus 3pt}%
  \setlength{\dblfloatsep}{3pt plus 2pt minus 2pt}%
  \setlength{\abovecaptionskip}{2pt}%
  \setlength{\belowcaptionskip}{2pt}%
}

\usepackage{listings}
\usepackage[normalem]{ulem}
\usepackage{xcolor}
\lstdefinestyle{promptstyle}{
    basicstyle=\ttfamily\footnotesize,
    breaklines=true,
    breakatwhitespace=false,
    columns=fullflexible,
    keepspaces=true,
    frame=single,
    framerule=0.4pt,
    xleftmargin=1em,
    xrightmargin=1em,
    aboveskip=0.8em,
    belowskip=0.8em
}
\usepackage{booktabs}
\usepackage{tabularx}
\usepackage{array}
\usepackage{multirow}
\usepackage{makecell}

\usepackage{newfloat}
\DeclareCaptionStyle{ruled}{labelfont=normalfont,labelsep=colon,strut=off} % DO NOT CHANGE THIS
\floatstyle{ruled}
\newfloat{listing}{tb}{lst}{}
\floatname{listing}{Listing}

\usepackage[most]{tcolorbox}

\usepackage[colorlinks=true,linkcolor=blue,citecolor=blue,urlcolor=blue]{hyperref}

\title{\fontsize{14}{19}\selectfont MemSIF: From Structured Interactions to Dual-Track Fact Memory for LLM Agents}

\author{
    YuFei Luo\textsuperscript{\rm 1},
    Xiucheng Xu\textsuperscript{\rm 2},
    Zhen Yang\textsuperscript{\rm 1}\corresponding
}
\affiliations{
    \textsuperscript{\rm 1}Beijing University of Posts and Telecommunications\\[2pt]
    \textsuperscript{\rm 2}University of Chinese Academy of Sciences\\[2pt]
    luoyf@bupt.edu.cn, xuxiucheng24s@ict.ac.cn, yangzhenyz@bupt.edu.cn
}

\begin{document}

\maketitle

\begin{abstract}
Long-term memory is critical for LLM agents operating over long-horizon
interactions. However, several persistent limitations of existing memory systems can be traced 
to two recurring misalignment patterns in long-term interaction settings: 
Temporal--Structural Misalignment (TSM) and Delayed Utility Manifestation (DUM). 
TSM arises when temporal proximity does not reliably align with topical or event-level relatedness, whereas DUM arises when write-time salience does not reliably predict future query utility.
To mitigate these misalignment patterns, we propose \textsc{MemSIF} (Memory with
Structured Interactions and Facts), a structured interaction-to-fact memory
framework. Structured Interaction Memory organizes raw interactions into
Topical Segments that preserve local topical coherence and Event Trajectories
that maintain cross-time event continuity. Dual-Track Fact Memory uses two
complementary tracks: CoreFact memory consolidates stable, schema-guided
information at write time, whereas ActiveFact memory forms facts on demand and
promotes those supported by multiple historical sources and recurring query
demand for reuse.
Experiments on LoCoMo and LongMemEval-S across five backbone LLMs show that
\textsc{MemSIF} achieves the highest Total ACC in all settings, outperforming
the strongest baseline by 2.29\%--8.79\% on LoCoMo and
2.87\%--6.15\% on LongMemEval-S.
These results support the effectiveness of combining Structured Interaction
Memory with Dual-Track Fact Memory to mitigate TSM and DUM.
Code is available at \url{https://github.com/luoyufeihaha/MemSIF}.
\end{abstract}

\section{Introduction}
\label{sec:intro}

LLM agents have evolved from single-turn response systems into systems capable of supporting long-horizon tasks \cite{xia2025agent0, singh2025agentic, schmidgall2025agent}. As interaction histories accumulate, processing them entirely within the context window becomes not only costly but also potentially detrimental to performance \cite{liu2024lost, du2025context, pollertlam2026beyond}. Building retrievable and updatable external memory is thus central to extending agent capabilities \cite{xiong2026memory,yu2026agentic,zhang2025survey,zheng2026lifelong}.

\begin{figure}[t]
    \centering
    \includegraphics[width=\columnwidth]{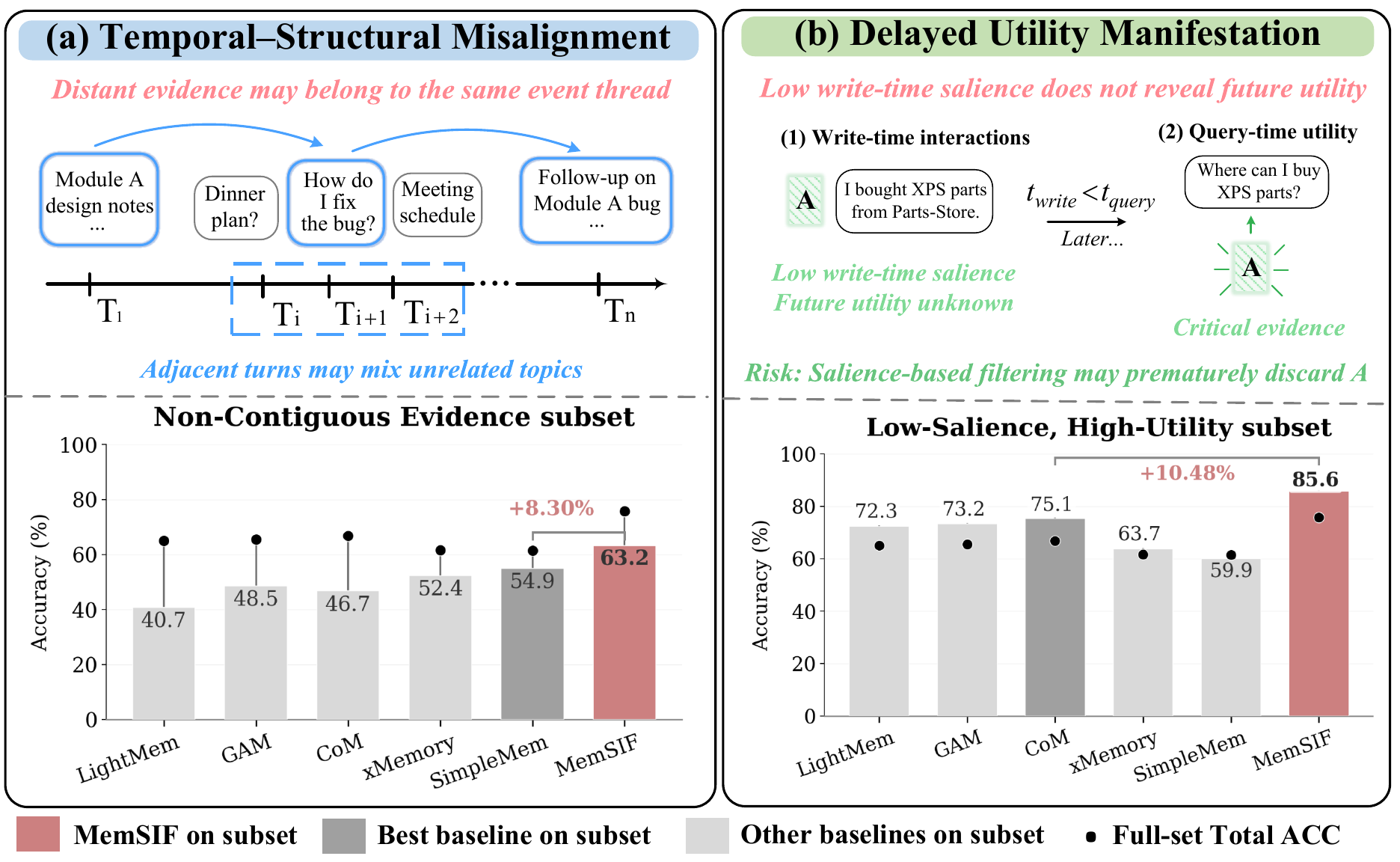}
\caption{
TSM and DUM as recurring mismatch patterns, with diagnostic-subset results on Qwen3-4B.
Subset construction details are provided in Appendix~\ref{app:phenomena}.
}
    \label{fig:motivation}
\end{figure}

Existing memory systems manage long-range context through summarization, retrieval, or structured indexing, 
reducing context overhead and improving access to historical evidence \cite{zhong2024memorybank,chhikara2025mem0,xu2026mem}. 
Recent studies identify two broad limitations in current long-term memory systems. 
The first concerns the organization of historical interactions, 
including topic fragmentation and difficulty in linking multi-hop evidence distributed across time \cite{zhang2026evolving,tan2025prospect,gutierrez2024hipporag}. 
The second concerns the assessment of information utility, 
including the premature removal of potentially useful information and the need to update stored information as user states evolve \cite{zhong2024memorybank,cham2026writepolicybench,huang2026rethinking}. 
These limitations are not isolated failures of specific systems. 
Rather, they reflect two recurring systematic misalignments in long-term memory design: Temporal--Structural Misalignment (TSM), 
in which temporal proximity does not reliably align with topical or event-level relatedness, and Delayed Utility Manifestation (DUM), 
in which write-time salience does not reliably predict the value of information for future queries.

Figure~\ref{fig:motivation}(a) illustrates the first mismatch pattern, TSM. Temporally adjacent conversation turns may address unrelated topics, whereas non-contiguous turns may represent different stages of the same evolving event or task. Consequently, predefined temporal boundaries may either group unrelated topics within the same memory unit or separate related evidence across multiple units, making it difficult to preserve both local topical coherence and cross-time event continuity. Existing methods seek to improve the organization of historical interactions using hierarchical memory structures, as in xMemory~\cite{hu2026beyond}, or self-contained memory entries, as in SimpleMem~\cite{liu2026simplemem}. However, because these approaches still operate over predefined memory units, evidence associated with the same event may remain dispersed across disconnected entries. The results on the Non-Contiguous Evidence (NCE) subset in Figure~\ref{fig:motivation}(a) further indicate that existing systems remain limited in their ability to link related evidence distributed across time.

Figure~\ref{fig:motivation}(b) illustrates the second mismatch pattern, DUM. The utility of stored information may change as subsequent interactions unfold and task demands change. Because future utility is unknown at write time, information that appears unimportant may become critical for answering later queries, making it difficult for a memory system to reliably determine which information should be retained or consolidated into reusable facts. Write-time consolidation methods, including Mem0~\cite{chhikara2025mem0} and MemoryOS~\cite{kang2025memory}, must select or compress information before future queries are known. They face a trade-off: aggressive filtering may discard information that later becomes critical, whereas conservative retention may preserve large amounts of redundant detail. Query-time methods, including CoM~\cite{xu2026chain} and GAM~\cite{yan2025general}, defer evidence selection until a query is observed and can recover initially low-salience information through evidence-chain organization or deep retrieval. However, the success of these query-time methods depends on retrieving the relevant raw interactions for each query. Because the recovered evidence is used only for the current query rather than persistently consolidated, the system must retrieve it again for subsequent queries, leaving future reuse vulnerable to retrieval misses. Results on the Low-Salience, High-Utility (LSHU) subset in Figure~\ref{fig:motivation}(b) further indicate that existing systems remain limited in their ability to retain, recover, reuse information whose value becomes apparent only after write time.

Existing long-term memory approaches typically optimize either interaction organization or fact construction independently. However, improving interaction organization cannot protect low-salience evidence from removal at write time, whereas mechanisms that account for utility assessment cannot repair evidence chains broken by temporal segmentation. To jointly mitigate TSM and DUM, we propose \textsc{MemSIF} (Memory with Structured Interactions and Facts), a structured interaction-to-fact memory framework that couples interaction organization with fact construction. \textsc{MemSIF} consists of two modules: Structured Interaction Memory and Dual-Track Fact Memory. To mitigate TSM, Structured Interaction Memory organizes interaction history into Topical Segments and Event Trajectories; the former preserves local topic coherence, while the latter maintains cross-time event continuity. Building on Structured Interaction Memory, Dual-Track Fact Memory mitigates DUM through two tracks: CoreFact and ActiveFact. CoreFact consolidates schema-guided stable information whose long-term value is clear at write time. ActiveFact forms candidate facts on demand when their utility becomes evident through queries. These candidates are promoted to reusable entries when supported by multiple historical sources and recurring query demand.

Experiments across five backbone LLMs show that \textsc{MemSIF} achieves the highest Total ACC in every setting on LoCoMo and LongMemEval-S, outperforming the strongest baseline by 2.29\%--8.79\% and 2.87\%--6.15\%, respectively. On the NCE and LSHU diagnostic subsets shown in Figure~\ref{fig:motivation}, \textsc{MemSIF} outperforms the strongest baseline by 8.30\% and 10.48\%, respectively. These targeted results further support its effectiveness in mitigating TSM and DUM. Our contributions are summarized as follows:
\begin{itemize}
\item We characterize TSM and DUM as two recurring misalignment patterns in long-term Agent memory, validate them through empirical diagnostics, and construct targeted diagnostic subsets for evaluation.
\item We propose \textsc{MemSIF}, a structured interaction-to-fact memory framework that jointly addresses TSM and DUM through structured interaction organization and dual-track fact memory.
\item We evaluate \textsc{MemSIF} across two long-term memory QA benchmarks and five backbone LLMs, demonstrating consistent Total ACC improvements under all settings, mechanistic evidence, and a favorable accuracy--efficiency trade-off.
\end{itemize}

\begin{figure*}[t]
    \centering
    \captionsetup{justification=raggedright, singlelinecheck=false}
    \includegraphics[width=\textwidth]{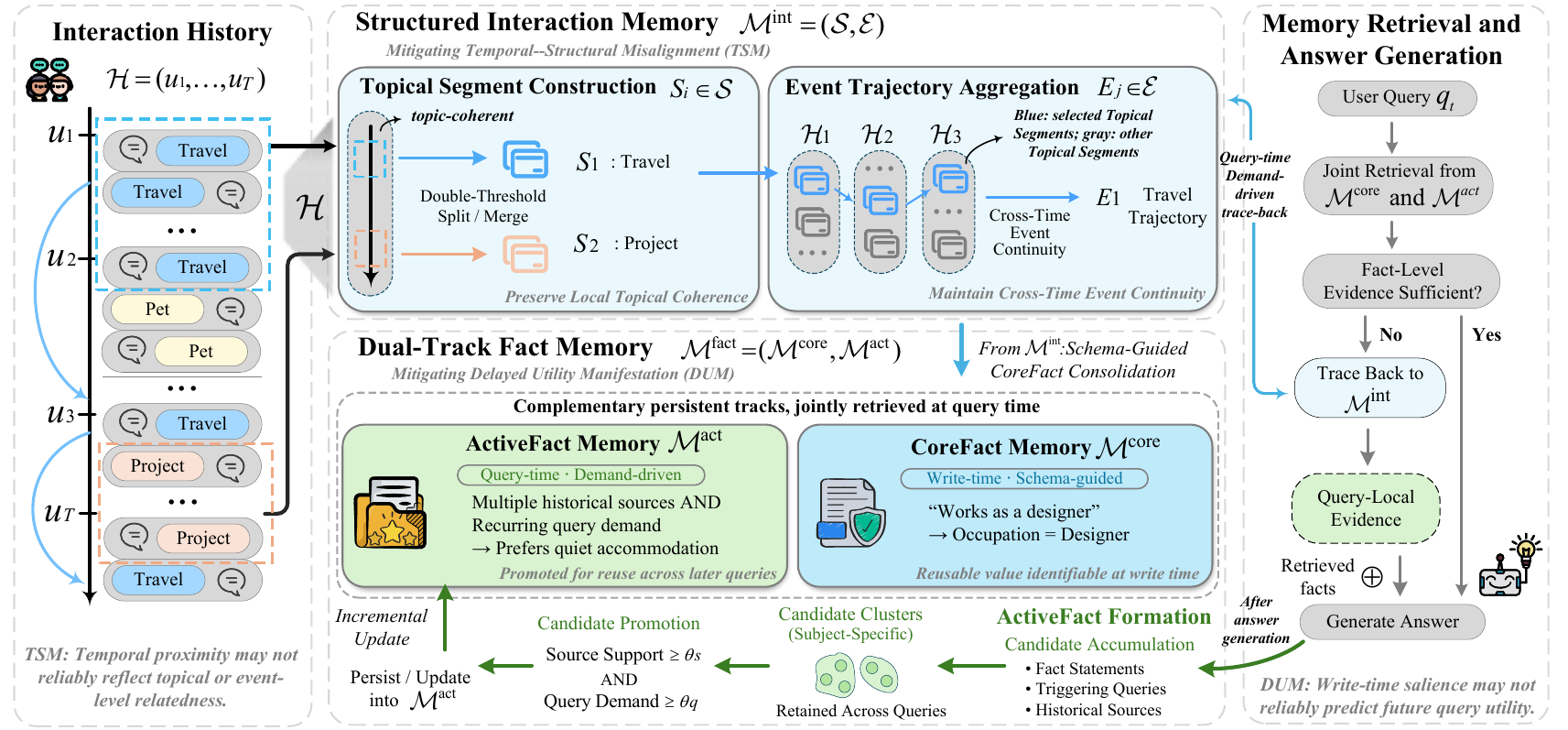}
    \caption{Overview of \textsc{MemSIF}. Structured Interaction Memory and Dual-Track Fact Memory jointly address TSM and DUM.}
    \label{fig:method}
\end{figure*}

\section{Related Work}

\subsection{Interaction Organization in Agent Memory}
\label{sec:interaction_org}

Interaction organization concerns how agent memory systems segment, index, and
retrieve historical interactions at different granularities
\cite{park2023generative,packer2023memgpt,zhong2024memorybank,lu2023memochat}.
Fine-grained approaches use messages, turns, or short spans as memory units
\cite{shinn2023reflexion,zhao2024expel}; for example, A-Mem
\cite{xu2026mem} constructs a MemoryNote for each turn. Coarse-grained
approaches organize larger contexts, such as the sliding-window entries in
SimpleMem \cite{liu2026simplemem} and the Pages and Sessions in MemoryOS
\cite{kang2025memory}. Fixed granularities make construction tractable
and retrieval efficient. Recent work introduces richer structures, including
xMemory's hierarchical memory components \cite{hu2026beyond}, SEEM's combination
of relational fact graphs with episodic event frames
\cite{lu2026structured}, Wu et al.'s \cite{wu2026gam} Event Progression Graph
and Topic Associative Network, and SAGE's \cite{wang2026sage} graph memory
refinement through reader--writer feedback.

Fixed-granularity units may nevertheless misalign with interaction structure.
Local topic shifts occur within a window or session, while related event
evidence recur across temporally separated units. Structured methods relax
fixed boundaries through hierarchical aggregation, event modeling,
consolidation, or graph evolution, yet they often treat interaction
organization and fact construction as separate design stages.
MemSIF instead jointly organizes interactions and constructs facts within a
unified framework, coupling contiguous Topical Segments with non-contiguous
Event Trajectories so that fact construction draws on both local coherence and
cross-time evidence continuity.

\subsection{Fact Construction in Agent Memory}

Fact construction concerns when agent memory systems consolidate
information into reusable facts and how those facts are updated as user states
and task demands evolve
\cite{zhang2025survey,huang2026rethinking}. Existing methods differ in when
they assess memory value and how they structure the resulting facts
\cite{cham2026writepolicybench,xiong2026memory}. 
Write-time approaches
\cite{modarressi2023ret} construct memory
before queries are known. Mem0 \cite{chhikara2025mem0} extracts and
consolidates salient information, while SimpleMem \cite{liu2026simplemem}
improves fact usability through self-contained memory formulation.
Query-time approaches defer
evidence selection and composition until a query is observed
\cite{qian2024memorag,sarthi2024raptor}; for example,
CoM \cite{xu2026chain} and GAM \cite{yan2025general} retrieve and combine
query-relevant evidence on demand. 
Structure-constrained methods guide construction with predefined memory
types or hierarchical organizations, such as Profile and Knowledge memories
in MemoryOS \cite{kang2025memory} and the component hierarchy in xMemory
\cite{hu2026beyond}.

Each strategy introduces a trade-off. Write-time methods
create compact reusable facts but may discard details whose utility appears later.
Query-time methods recover task-relevant evidence but primarily construct
query-specific contexts rather than persist reusable facts.
Structure-constrained methods improve consistency but may not accommodate
future needs beyond their predefined structures.
MemSIF addresses these trade-offs through Dual-Track Fact Memory. CoreFact memory
consolidates write-time facts, while ActiveFact memory forms facts on demand when their utility
emerges only later and traces back to Structured Interaction Memory when
fact-level evidence is insufficient. This design couples interaction
organization with fact construction, making the joint treatment of TSM and DUM
its central focus.

\section{Method}
\subsection{Overview}

As shown in Figure~\ref{fig:method}, \textsc{MemSIF} is a structured interaction-to-fact memory framework for
long-term agent memory. Given an interaction history $\mathcal{H}$, MemSIF
constructs Structured Interaction Memory
$\mathcal{M}^{\mathrm{int}}=(\mathcal{S},\mathcal{E})$ and maintains Dual-Track Fact Memory
$\mathcal{M}^{\mathrm{fact}}=(\mathcal{M}^{\mathrm{core}},
\mathcal{M}^{\mathrm{act}})$.
Structured Interaction Memory organizes $\mathcal{H}$ into Topical Segments
$S_i\in\mathcal{S}$ and
Event Trajectories $E_j\in\mathcal{E}$ to preserve local topic boundaries and cross-time event
continuity, addressing TSM. Dual-Track Fact Memory mitigates DUM through two complementary tracks. CoreFact memory consolidates stable, schema-guided information whose reusable value can be identified at write time, whereas ActiveFact memory forms candidate facts on demand when their utility becomes evident through subsequent queries. Both tracks ultimately maintain persistent, updatable facts but differ in the timing and evidence used for consolidation. Throughout this section, a fact refers to a normalized, provenance-linked memory record rather than an externally verified proposition.

\subsection{Structured Interaction Memory}
\label{sec:structured_interaction_memory}

To mitigate TSM without relying on predefined temporal boundaries, Structured Interaction Memory constructs two complementary representations from raw interaction history: Topical Segments and Event Trajectories.

Given an interaction history
$\mathcal{H}=(u_1,u_2,\ldots,u_T)$, where each $u_t$ is a single message,
MemSIF constructs Structured Interaction Memory
$\mathcal{M}^{\mathrm{int}}=(\mathcal{S},\mathcal{E})$. A
Topical Segment $S_i\in\mathcal{S}$ is a temporally contiguous,
topic-coherent interaction unit. An Event Trajectory
$E_j\in\mathcal{E}$ associates non-contiguous Topical Segments
belonging to the same evolving event or task. These two structures jointly organize $\mathcal{H}$ for downstream fact construction.
MemSIF uses a shared matching function for segment construction and trajectory aggregation:
\begin{equation}
\label{eq:interaction_matching}
\phi(A,B)=
\alpha s_{\mathrm{sem}}(A,B)
+(1-\alpha)J(\mathcal{K}_A,\mathcal{K}_B),
\end{equation}
where $A$ and $B$ can be messages, Topical Segments, or Event Trajectories. 
The weight $\alpha$ balances semantic similarity and
entity overlap. The semantic score $s_{\mathrm{sem}}$ is normalized cosine
similarity and $J$ computes Jaccard overlap between key-entity sets. For composite
units such as segments and trajectories, text embeddings are mean-pooled from
constituent messages and entity sets are unioned. After each append or
assignment, the affected unit's embedding and entity set are updated
accordingly.

\paragraph{Topical Segment Construction.}
MemSIF processes messages chronologically. The first message initializes
$S_1$. For each subsequent message $u_t$, MemSIF compares it with the current
segment $S_i$ using $\phi(u_t,S_i)$ and applies a double-threshold rule with
$\tau_{\mathrm{split}}<\tau_{\mathrm{merge}}$. If
$\phi(u_t,S_i)\geq\tau_{\mathrm{merge}}$, the message is appended to $S_i$; if
$\phi(u_t,S_i)\leq\tau_{\mathrm{split}}$, it starts a new segment. Scores
between thresholds invoke an LLM to determine whether the message
continues the topic; if it does, the message is appended to $S_i$, otherwise a
new segment is created. This design delegates only ambiguous boundary
decisions to the LLM, keeping construction efficient while preserving
topic coherence.

\paragraph{Event Trajectory Aggregation.}
MemSIF next processes Topical Segments chronologically. For each $S_i$, it
retrieves the top-$K$ candidate trajectories according to $\phi(S_i,E_j)$ and
uses an LLM to assess event identity, task continuity, and state
consistency. If multiple trajectories are compatible, MemSIF selects the
highest-scoring one and appends $S_i$ to it. If no trajectory is compatible, $S_i$ initializes a new
trajectory.

\subsection{Schema-Guided CoreFact Consolidation}
\label{sec:corefact}

CoreFact memory $\mathcal{M}^{\mathrm{core}}$ consolidates information whose
reusable value can be identified at write time. Converting Structured Interaction Memory contents into fact memory can introduce
redundancy and increase retrieval cost; MemSIF adopts a configurable CoreFact
schema $\mathcal{Y}^{\mathrm{core}}$ that specifies compact, reusable fact
types for the target setting, such as user preferences and task states. The
schema is specified before memory construction and fixed within each setting.

An LLM first extracts schema-eligible candidate facts from Topical Segments.
Event Trajectories then provide cross-time evidence to supplement or reconcile
facts that span multiple segments. Each CoreFact entry is represented as
$f^{\mathrm{core}}=\langle\text{subject},\text{type},\text{stmt},\text{meta}\rangle$, where
$\text{subject}$ is the described entity,
$\text{type}\in\mathcal{Y}^{\mathrm{core}}$ is its schema category, and
$\text{stmt}$ is the normalized statement. The metadata records the timestamp,
source provenance, and validity status.

Before writing, MemSIF retrieves CoreFact entries using subject,
type, and semantic similarity. A schema-constrained LLM selects one of four
operations. \textsc{Add} creates a new fact; \textsc{Update} integrates
compatible new evidence into an existing fact; \textsc{Merge} consolidates
duplicate or near-duplicate entries; and \textsc{Supersede} marks an outdated
fact as replaced while retaining it for historical queries. Information outside
$\mathcal{Y}^{\mathrm{core}}$ or uncertain at write time remains in Structured
Interaction Memory as recoverable evidence for ActiveFact formation.

\begin{table*}[t]
\centering
\resizebox{\textwidth}{!}{%
\small\setlength{\tabcolsep}{2.8pt}
\begin{tabular}{l|ccccc|ccccc|ccccc}
\toprule
\multirow{2}{*}{Method}
  & \multicolumn{5}{c|}{Qwen3-4B}
  & \multicolumn{5}{c|}{Qwen3-32B}
  & \multicolumn{5}{c}{DeepSeek-v4-pro} \\
\cmidrule(lr){2-6} \cmidrule(lr){7-11} \cmidrule(lr){12-16}
 & S-hop & M-hop & Temp & Kno & Total
 & S-hop & M-hop & Temp & Kno & Total
 & S-hop & M-hop & Temp & Kno & Total \\
\midrule
Full-Context & 78.94 & 48.06 & 38.77 & 45.74 & 62.89
  & 85.04 & 65.09 & 59.41 & \underline{59.85} & 74.47
  & 85.48 & 75.99 & 78.53 & 63.48 & 80.92 \\
Naive RAG    & 69.91 & 37.59 & 40.00 & 29.17 & 55.25
  & 77.85 & 46.90 & 53.06 & 39.04 & 64.63
  & 80.45 & 50.34 & 70.98 & 51.76 & 71.18 \\
Mem0         & 59.48 & 46.10 & 26.25 & 35.42 & 48.64
  & 57.56 & 45.46 & 47.69 & 44.38 & 52.47
  & 65.48 & 48.82 & 51.89 & 47.33 & 58.47 \\
MemoryOS    & 64.93 & 43.26 & 23.13 & 41.67 & 50.84
  & 71.77 & 48.91 & 33.94 & 41.64 & 57.86
  & 81.76 & 55.71 & 38.66 & 47.44 & 65.86 \\
MemGAS       & 70.58 & 45.01 & 33.94 & 40.59 & 56.43
  & 74.15 & 50.32 & 41.71 & 51.01 & 61.62
  & 82.63 & 56.07 & 46.48 & 56.84 & 68.62 \\
LightMem     & 76.54 & 53.19 & \underline{53.12} & 37.50 & 64.98
  & 83.89 & 61.70 & \underline{68.13} & 43.75 & 74.06
  & 83.01 & \textbf{80.47} & 82.05 & \underline{63.85} & 81.15 \\
GAM          & 77.73 & 55.35 & 49.39 & 41.15 & 65.48
  & 87.15 & \underline{68.52} & 64.31 & 58.85 & \underline{77.21}
  & 88.43 & 73.09 & \textbf{83.41} & 62.79 & 82.97 \\
CoM          & \underline{80.54} & 59.39 & 44.34 & 43.10 & \underline{66.83}
  & 86.02 & 66.67 & 65.00 & 56.25 & 76.26
  & 84.42 & 69.15 & 77.88 & 55.21 & 78.44 \\
xMemory      & 75.36 & 56.03 & 34.38 & \underline{45.83} & 61.48
  & 80.09 & 50.10 & 55.26 & 52.01 & 67.70
  & 81.72 & 66.66 & 70.41 & 62.07 & 75.38 \\
SimpleMem    & 73.67 & \underline{59.89} & 35.18 & 44.76 & 61.36
  & \underline{87.23} & 63.09 & 57.29 & 55.18 & 74.61
  & \underline{90.11} & 75.49 & 80.01 & 60.02 & \underline{83.45} \\
\midrule
MemSIF       & \textbf{85.44} & \textbf{61.96} & \textbf{69.49} & \textbf{50.38} & \textbf{75.62}$^\dagger$
  & \textbf{91.20} & \textbf{68.79} & \textbf{79.13} & \textbf{65.62} & \textbf{82.99}$^\dagger$
  & \textbf{91.79} & \underline{78.29} & \underline{82.59} & \textbf{65.30} & \textbf{85.74}$^\dagger$ \\
\bottomrule
\end{tabular}%
}
\caption{Main experimental results on LoCoMo. Results report mean ACC
(\%) over three runs. \textbf{Bold} and \underline{underlining}
denote the best and second-best results in each column.
For Total ACC, $^\dagger$ shows the 95\% paired-bootstrap
confidence interval for MemSIF's improvement over the strongest baseline
under the same backbone excludes zero.}
\label{tab:main_results}
\end{table*}

\begin{table*}[t]
\centering
\resizebox{\textwidth}{!}{%
\small\setlength{\tabcolsep}{3pt}
\begin{tabular}{l|ccccc|ccccc|ccccc}
\toprule
\multirow{2}{*}{Method}
  & \multicolumn{5}{c|}{Qwen3-4B}
  & \multicolumn{5}{c|}{Qwen3-32B}
  & \multicolumn{5}{c}{DeepSeek-v4-pro} \\
\cmidrule(lr){2-6} \cmidrule(lr){7-11} \cmidrule(lr){12-16}
 & S-hop & M-hop & Temp & Kno & Total
 & S-hop & M-hop & Temp & Kno & Total
 & S-hop & M-hop & Temp & Kno & Total \\
\midrule
Full-Context & 44.19 & 29.95 & 24.47 & 35.35 & 33.84
  & 52.83 & 40.93 & 39.99 & 59.24 & 47.28
  & 61.70 & 47.80 & 46.71 & 69.18 & 55.22 \\
Naive RAG    & 62.25 & 44.56 & 23.46 & 51.25 & 45.53
  & 73.97 & 54.06 & 45.56 & \underline{75.29} & 61.37
  & 80.53 & 59.32 & 53.83 & 70.01 & 66.24 \\
LightMem     & \underline{75.59} & 47.23 & \underline{56.02} & 69.59 & 62.08
  & \underline{81.17} & 53.06 & 62.11 & 74.28 & 67.73
  & 84.50 & 56.83 & \underline{68.71} & \underline{79.36} & 72.32 \\
GAM          & 70.52 & \underline{51.78} & 35.47 & 61.02 & 54.77
  & 76.87 & 56.25 & 43.22 & 69.33 & 61.31
  & 79.50 & 60.50 & 49.03 & 73.62 & 65.47 \\
CoM          & 74.75 & 51.28 & 54.48 & \underline{71.41} & \underline{62.72}
  & 80.52 & \underline{60.13} & 62.77 & 72.87 & \underline{69.30}
  & 83.35 & 63.50 & 62.21 & 77.56 & 71.64 \\
SimpleMem    & 70.98 & 50.22 & 52.28 & 68.27 & 60.17
  & 78.40 & 58.83 & \underline{65.10} & 56.11 & 66.35
  & \underline{84.57} & \underline{65.86} & 67.45 & 76.27 & \underline{73.86} \\
\midrule
MemSIF       & \textbf{79.76} & \textbf{57.72} & \textbf{63.61} & \textbf{74.22} & \textbf{68.87}$^\dagger$
  & \textbf{83.41} & \textbf{64.20} & \textbf{69.15} & \textbf{78.87} & \textbf{73.92}$^\dagger$
  & \textbf{86.54} & \textbf{67.69} & \textbf{70.60} & \textbf{82.30} & \textbf{76.73}$^\dagger$ \\
\bottomrule
\end{tabular}%
}
\caption{Main experimental results on LongMemEval-S. Results report mean ACC
(\%) over three runs. \textbf{Bold} and \underline{underlining}
denote the best and second-best results in each column.
For Total ACC, $^\dagger$ shows the 95\% paired-bootstrap
confidence interval for MemSIF's improvement over the strongest baseline
under the same backbone excludes zero.}
\label{tab:main_results_lmes}
\end{table*}

\subsection{Query-Driven ActiveFact Formation}
\label{sec:activefact}

ActiveFact memory $\mathcal{M}^{\mathrm{act}}$ targets information whose
reusable value is uncertain at write time and may become evident through
subsequent queries. It operates through four stages: query-local extraction,
post-answer candidate accumulation, candidate promotion, and persistent
maintenance. Query-local evidence can support the current answer without
immediate consolidation, whereas candidates supported by multiple historical
sources and recurring query demand are promoted into persistent entries for direct reuse, rather than requiring repeated reconstruction from raw history.

\paragraph{Query-Local Extraction and Candidate Accumulation.}
When the retrieved fact-level evidence is insufficient for a query $q_t$,
MemSIF retrieves relevant Topical Segments and Event Trajectories and extracts
query-local evidence. Each evidence entry contains a normalized statement,
subject, and source provenance. It supports the generation of the current
answer $a_t$ but is not written directly into ActiveFact memory.

Only after $a_t$ has been generated does each evidence entry become eligible
for candidate accumulation. MemSIF organizes candidate facts into
subject-specific clusters
$C_k=(s_k,\mathcal{X}_k,\mathcal{Q}_k,\mathcal{R}_k)$, where $s_k$ denotes the
normalized subject, $\mathcal{X}_k$ contains deduplicated fact statements,
$\mathcal{Q}_k$ contains distinct triggering queries after near-duplicate
removal, and $\mathcal{R}_k$ contains identifiers of distinct historical
interaction units in the evidence provenance. An evidence entry is
merged into an existing cluster when its subject matches $s_k$ and its
statement is compatible with the accumulated statements in
$\mathcal{X}_k$; otherwise, it initializes a new cluster.

Candidate updates use only the extracted statement, its source provenance, and
the triggering query $q_t$. Query-local evidence is cleared after each answer,
whereas candidate clusters are retained across queries within the same
interaction history, allowing historical support and query demand to accumulate
over time.

\paragraph{Candidate Promotion and Maintenance.}
As candidate clusters accumulate, MemSIF evaluates each cluster using two
signals: source support and query demand:
\begin{align}
\mathrm{Score}_{\mathrm{src}}(C_k)
&=
\left(1-1/|\mathcal{R}_k|\right)
\cdot \mathrm{Coh}(\mathcal{X}_k), \\
\mathrm{Score}_{\mathrm{qry}}(C_k)
&=
\left(1-1/|\mathcal{Q}_k|\right)
\cdot \mathrm{Coh}(\mathcal{Q}_k).
\end{align}
$|\mathcal{R}_k|$ and $|\mathcal{Q}_k|$ denote the numbers of distinct
historical sources and retained triggering queries, respectively.
$\mathrm{Coh}(\cdot)$ denotes the average pairwise normalized cosine similarity
in a set; for a singleton set, $\mathrm{Coh}(\cdot)=1$.
The source score captures the amount and consistency of historical support,
whereas the query score captures repeated and semantically coherent demand.
The promotion decision is defined as:
\begin{equation}
\scalebox{0.95}{$
P(C_k)=
\mathbf{1}\left[
\mathrm{Score}_{\mathrm{src}}(C_k)\geq\theta_s
\land
\mathrm{Score}_{\mathrm{qry}}(C_k)\geq\theta_q
\right].
$}
\end{equation}
A candidate is promoted only when it is both historically supported and
repeatedly demanded. When $P(C_k)=1$, MemSIF normalizes the cluster into a
promoted ActiveFact entry $\tilde f_k^{\mathrm{act}} = \langle\text{subject},\text{stmt},\text{meta}\rangle$, where $\text{subject}=s_k$, $\text{stmt}$ is the normalized statement synthesized from $\mathcal{X}_k$, and $\text{meta}$ records provenance,
timestamps, and triggering-query references. The resulting entry is stored in
ActiveFact memory $\mathcal{M}^{\mathrm{act}}$. In continual deployment,
compatible entries are updated as subsequent queries surface additional
evidence from the evolving interaction history.

\subsection{Memory Retrieval and Answer Generation}
\label{sec:retrieval_generation}

For each query $q_t$, MemSIF retrieves relevant CoreFact entries from
$\mathcal{M}^{\mathrm{core}}$ and persistent ActiveFact entries from
$\mathcal{M}^{\mathrm{act}}$ by semantic similarity. An LLM-based sufficiency
checker determines whether the retrieved fact-level evidence covers the
information required by $q_t$. When sufficient, the retrieved entries form the
answer context; otherwise, MemSIF retrieves query-local evidence from
Structured Interaction Memory $\mathcal{M}^{\mathrm{int}}$. The answer context then combines the retrieved fact entries with the
query-local evidence rather than directly including the raw interaction units.
The LLM generates $a_t$ from the answer context. After generation, the
query-local evidence enters the ActiveFact pipeline. Appendix~\ref{app:case_studies} illustrates MemSIF's behavior under TSM and DUM through concrete case studies.

\begin{table*}[t]
\centering
\resizebox{\textwidth}{!}{%
\setlength{\tabcolsep}{4pt}
\begin{tabular}{l|ccccc|ccccc}
\toprule
\multirow{2}{*}{Method}
  & \multicolumn{5}{c|}{LoCoMo}
  & \multicolumn{5}{c}{LongMemEval-S} \\
\cmidrule(lr){2-6} \cmidrule(lr){7-11}
 & S-hop & M-hop & Temp & Kno & Total
 & S-hop & M-hop & Temp & Kno & Total \\
\midrule
MemSIF & \textbf{85.44} & \textbf{61.96} & \textbf{69.49} & \textbf{50.38} & \textbf{75.62}
  & \textbf{79.76} & \textbf{57.72} & \textbf{63.61} & \textbf{74.22} & \textbf{68.87} \\
\midrule
w/o TS
  & \dvA{79.83}{-5.61} & \dvA{53.63}{-8.33} & \dvA{62.77}{-6.72} & \dvA{47.41}{-2.97} & \dvA{69.46}{-6.16}
  & \dvA{73.18}{-6.58} & \dvA{48.25}{-9.47} & \dvA{57.78}{-5.83} & \dvA{68.88}{-5.34} & \dvA{61.94}{-6.93} \\
w/o ET
  & \dvA{81.33}{-4.11} & \dvA{56.18}{-5.78} & \dvA{59.13}{\textbf{-10.36}} & \dvA{49.87}{-0.51} & \dvA{70.14}{-5.48}
  & \dvA{75.83}{-3.93} & \dvA{52.55}{-5.17} & \dvA{56.73}{-6.88} & \dvA{69.91}{-4.31} & \dvA{63.77}{-5.10} \\
w/o CF
  & \dvA{75.17}{\textbf{-10.27}} & \dvA{56.17}{-5.79} & \dvA{66.86}{-2.63} & \dvA{39.22}{\textbf{-11.16}} & \dvA{67.72}{-7.90}
  & \dvA{65.48}{\textbf{-14.28}} & \dvA{49.25}{-8.47} & \dvA{55.93}{-7.68} & \dvA{61.43}{\textbf{-12.79}} & \dvA{56.10}{\textbf{-12.77}} \\
w/o AF
  & \dvA{78.27}{-7.17} & \dvA{48.73}{\textbf{-13.23}} & \dvA{63.11}{-6.38} & \dvA{41.77}{-8.61} & \dvA{67.43}{\textbf{-8.19}}
  & \dvA{72.33}{-7.43} & \dvA{45.34}{\textbf{-12.38}} & \dvA{55.72}{\textbf{-7.89}} & \dvA{64.68}{-9.54} & \dvA{59.72}{-9.15} \\
\bottomrule
\end{tabular}%
}
\caption{Ablation study of MemSIF on Qwen3-4B. Each cell shows absolute ACC (\%) with $\Delta$ from MemSIF Full. \textbf{Bold} $\Delta$ marks the largest drop per column.}
\label{tab:ablation}
\end{table*}

\begin{figure*}[t]
\centering
\includegraphics[width=\textwidth]{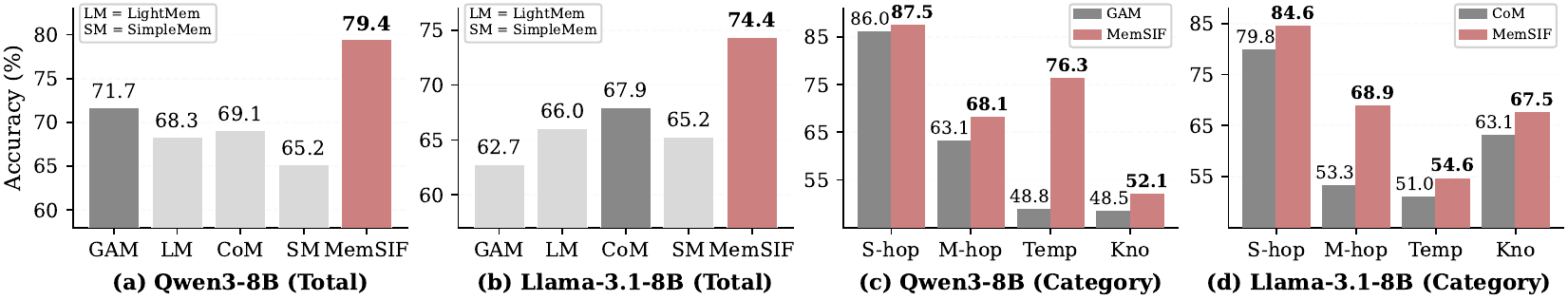}
\caption{Backbone generalization on LoCoMo. Panels (a,b): Total ACC of representative baselines on Qwen3-8B and Llama-3.1-8B-Instruct. Panels (c,d): MemSIF vs. strongest per-backbone baseline across question categories. Full results in Appendix~\ref{app:supplementary_results}.}
\label{fig:backbone_generalization}
\end{figure*}

\section{Experiments}
\subsection{Experimental Setup}
\label{sec:experimental_setup}

\paragraph{Benchmark Dataset.}
We evaluate MemSIF on two long-term memory QA benchmarks, LoCoMo
\cite{maharana2024evaluating} and LongMemEval-S
\cite{wu2024longmemeval}. LoCoMo contains 10 two-party conversations, each
spanning approximately 27 sessions and averaging 16.6K tokens. We exclude adversarial questions and retain 1,540
evidence-grounded samples. LongMemEval-S contains 500 independent samples with
interaction histories averaging over 100K tokens. After excluding 30
abstention questions, 470
samples remain for evaluation. Dataset statistics are provided in
Appendix~\ref{app:dataset_stats}.

\paragraph{Diagnostic Subsets.}
Two diagnostic subsets are constructed from LoCoMo gold evidence.
Non-Contiguous Evidence (NCE) subset selects questions whose gold evidence
spans widely separated turns, measured by Temporal Evidence Dispersion
($\mathrm{TED}\geq0.3$). The Low-Salience, High-Utility (LSHU) subset selects
questions with evidence of low write-time salience yet high
query-time utility ($S(e)\leq2$, $U(e,q)=3$). Full construction
criteria and empirical analyses of TSM and DUM are provided in
Appendix~\ref{app:phenomena}.

\paragraph{Baselines.}
We compare MemSIF with 10 baselines: Full-Context and Naive~RAG as basic bounds;
Mem0~\cite{chhikara2025mem0}, MemoryOS~\cite{kang2025memory}, MemGAS~\cite{xu2025single}, xMemory~\cite{hu2026beyond},
and SimpleMem~\cite{liu2026simplemem} as write-time consolidation methods; and
LightMem~\cite{fang2025lightmem}, GAM~\cite{yan2025general}, and CoM~\cite{xu2026chain} as query-time
composition methods.

\paragraph{Evaluation Metric.}
We use Accuracy (ACC) as the primary metric. A GPT-4o
\cite{hurst2024gpt} judge labels an answer as correct if it directly addresses
the question and is semantically consistent with the essential reference
information. We validate the evaluator on 440 answers sampled from both
datasets and all systems under Qwen3-4B. Two annotators, blinded to system
identity and GPT-4o labels, independently evaluate each sample using the same
criteria. GPT-4o achieves 93.0\% agreement with the adjudicated human labels
(Cohen's $\kappa$=0.81); full protocols and analyses are provided in
Appendix~\ref{app:judge_validation}.

\paragraph{Implementation Details.}
The main experiments use Qwen3-4B, Qwen3-32B~\cite{yang2025qwen3}, and
DeepSeek-v4-pro~\cite{xu2026deepseek} (all in non-thinking mode) on both
datasets. We evaluate Qwen3-8B~\cite{yang2025qwen3} (non-thinking)
and Llama-3.1-8B-Instruct~\cite{grattafiori2024llama} on LoCoMo. We use
Qwen3-Embedding-8B~\cite{zhang2025qwen3} as the embedding backbone.
MemSIF uses $\alpha=0.8$ for the interaction matching function,
$\tau_{\mathrm{split}}=0.325$ and $\tau_{\mathrm{merge}}=0.60$ for Topical
Segment boundaries, Top-$K=3$ for Event Trajectory aggregation, and
$\theta_s=\theta_q=0.45$ for ActiveFact promotion.
To quantify the uncertainty of reported gains, we perform paired
bootstrap resampling with 10,000 resamples between MemSIF and the
strongest baseline in each dataset--backbone setting. We report the
95\% confidence interval of $\Delta\mathrm{ACC} =
\mathrm{ACC}_{\textsc{MemSIF}} -
\mathrm{ACC}_{\text{baseline}}$. Full methodology is provided in
Appendix~\ref{app:bootstrap}.
Appendix~\ref{app:implementation_details} provides implementation
details.

\begin{table}[t]
\centering
{\small
\setlength{\tabcolsep}{3.2pt}
\begin{tabular}{lcccc}
\toprule
Variant & Total & LSHU & Tokens/q & Time/q (s) \\
\midrule
w/o AF & 67.43 & 74.45 & 1,935 & 1.14 \\
Query-local only & 73.09 & 81.28 & 3,364 & 1.53 \\
Full MemSIF & \textbf{75.62} & \textbf{85.69} & 3,052 & 1.40 \\
\bottomrule
\end{tabular}
}
\caption{ActiveFact formation analysis on LoCoMo with Qwen3-4B. Query-local: query-local extraction only.}
\label{tab:activefact_formation}
\end{table}

\subsection{Main Results}
\label{sec:main_results}
Tables~\ref{tab:main_results} and~\ref{tab:main_results_lmes} report ACC for
single-hop (S-hop), multi-hop (M-hop), temporal (Temp), and knowledge (Kno)
questions. MemSIF achieves the highest Total ACC across both datasets and all
three main backbones, supporting its effectiveness across interaction
lengths and model capacities. We examine the contributions of its individual
components below.

On LoCoMo, MemSIF outperforms the strongest baseline by 8.79\%, 5.78\%, and
2.29\% under Qwen3-4B, Qwen3-32B, and DeepSeek-v4-pro, respectively. The
largest gain occurs under Qwen3-4B, suggesting that MemSIF is particularly
beneficial when backbone capacity is limited. The largest category-level gains
under Qwen3-4B and Qwen3-32B occur on Temp (16.37\% and 11.00\%), consistent
with the role of Event Trajectories in recovering temporally distributed
evidence. Under DeepSeek-v4-pro, MemSIF trails LightMem by 2.18\% on M-hop
and GAM by 0.82\% on Temp, indicating that MemSIF's gains are not uniform
across all categories under the strongest backbone. MemSIF nonetheless
achieves the highest Total ACC under all three backbones.

On LongMemEval-S, MemSIF exceeds the strongest baseline by 6.15\%, 4.62\%, and
2.87\% under Qwen3-4B, Qwen3-32B, and DeepSeek-v4-pro, respectively. Unlike
LoCoMo, where the largest gains concentrate on Temp, MemSIF leads all question
categories across the three backbones on this benchmark. Since each query is
paired with an independent history exceeding 100K tokens, evidence access
may place greater emphasis on long-context evidence filtering. The broad
category-level gains are consistent with the role of Dual-Track Fact Memory:
CoreFact memory keeps reusable information accessible, while ActiveFact memory
forms facts from evidence whose utility emerges at query time.

Figure~\ref{fig:backbone_generalization} further evaluates Qwen3-8B and
Llama-3.1-8B-Instruct on LoCoMo. MemSIF outperforms the strongest baseline by
7.69\% and 6.47\%, respectively, with category-level patterns consistent with
the three main backbones. These results further support MemSIF's generalization
across model scales and families. Complete numerical results are provided in
Appendix~\ref{app:supplementary_results}.

Paired bootstrap analysis further supports the reliability of these gains:
the 95\% confidence intervals for Total ACC improvement over the
strongest baseline exclude zero in all six dataset--backbone settings.
This includes smaller gains under DeepSeek-v4-pro on LoCoMo
(+2.29 \%, 95\% CI [0.45, 5.03]) and LongMemEval-S
(+2.87 \%, 95\% CI [0.32, 6.44]). Full confidence intervals appear
in Appendix~\ref{app:bootstrap}.

\subsection{Ablation and Mechanism Analysis}
\label{sec:ablation}
To understand how each design contributes to MemSIF's performance, we
conduct component ablation, diagnostic subset analysis, and sensitivity
studies. We ablate four components on both datasets with Qwen3-4B
(Table~\ref{tab:ablation}). \textbf{w/o TS} constructs Event Trajectories
directly from raw messages. \textbf{w/o ET} uses Topical Segments alone.
\textbf{w/o CF} removes CoreFact memory, relying solely on
ActiveFact memory. \textbf{w/o AF} disables ActiveFact memory,
keeping only CoreFact memory.

Removing Event Trajectories causes the largest Temp drop (10.36\% on LoCoMo),
consistent with its role in cross-time event continuity; removing
Topical Segments impairs M-hop, which requires topic-coherent units to
assemble evidence across segments. Removing CoreFact memory is most
consequential on LongMemEval-S (12.77\% Total decrease), where maintaining
compact, reusable facts becomes important under 100K+ token
contexts; removing ActiveFact memory causes the largest M-hop losses
(13.23\% and 12.38\%), reflecting its role in recovering evidence whose
utility emerges at query time. No component dominates all patterns; TSM and DUM require complementary solutions.

\begin{figure}[t]
\centering
\includegraphics[width=\columnwidth]{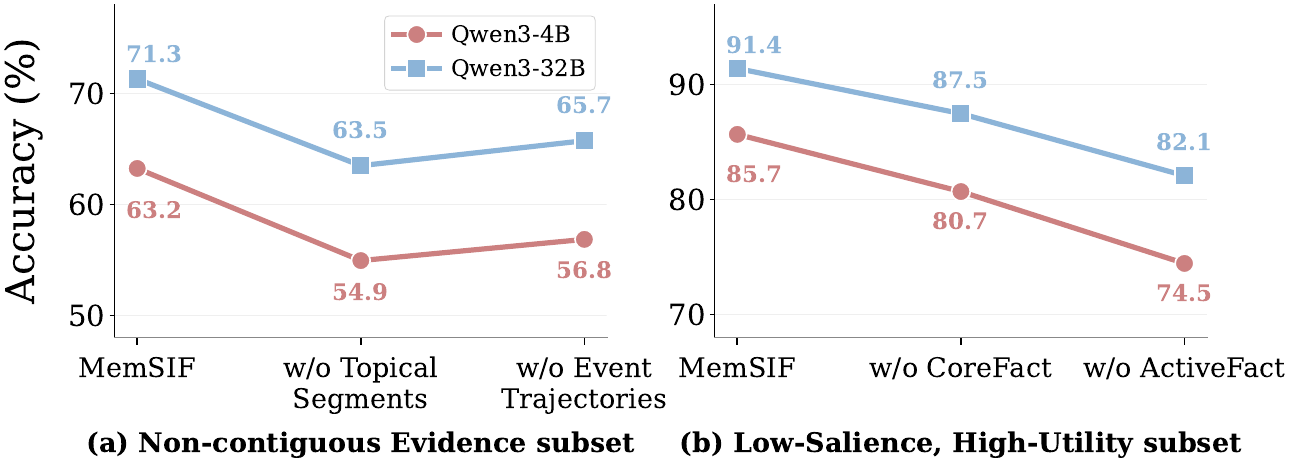}
\caption{Diagnostic-subset ablation results under Qwen3-4B and Qwen3-32B.}
\label{fig:ablation_linechart}
\end{figure}

To directly test how MemSIF handles TSM and DUM, we conduct ablation on two
diagnostic subsets (Figure~\ref{fig:ablation_linechart}). On the NCE subset,
removing Event Trajectories or Topical Segments causes the largest drops
(5.56\%--8.30\%), showing that interaction structuring is central to
mitigating TSM. On the LSHU subset, removing ActiveFact memory produces the
largest declines (9.34\%--11.24\%), while removing CoreFact memory causes
moderate drops (3.93\%--4.98\%). The larger drop from removing ActiveFact
memory identifies query-driven formation as the primary mechanism on LSHU,
while the smaller CoreFact memory drop indicates a complementary contribution
from reusable write-time facts.

Table~\ref{tab:activefact_formation} reports ActiveFact formation analysis.
Query-local only achieves 73.09\% Total ACC, already exceeding the strongest
LoCoMo baseline (CoM, 66.83\%), showing that on-demand evidence extraction is
effective even without persistent candidate accumulation. Full MemSIF further
improves LSHU ACC by 4.41\% over Query-local only while reducing token
usage and runtime, as persistent ActiveFact entries retained from earlier
queries can be reused without trace-back to Structured Interaction
Memory.

MemSIF is robust to its construction-time hyperparameters. Varying the
matching weight changes Total ACC by only 2.83\%, indicating that performance
does not depend on a finely tuned value. Combining semantic similarity with
entity overlap (75.60\%) outperforms either signal alone, showing that the two
signals provide complementary evidence. For segment construction, the default
double-threshold setting outperforms single-threshold alternatives, and varying
the gap width from 0.075 to 0.475 keeps Total ACC within the range
71.13\%--75.62\%, with the default setting performing best. Complete results
are in Appendix~\ref{app:sensitivity}.

\begin{figure}[t]
\captionsetup{justification=raggedright,singlelinecheck=false}
\centering
\includegraphics[width=\columnwidth]{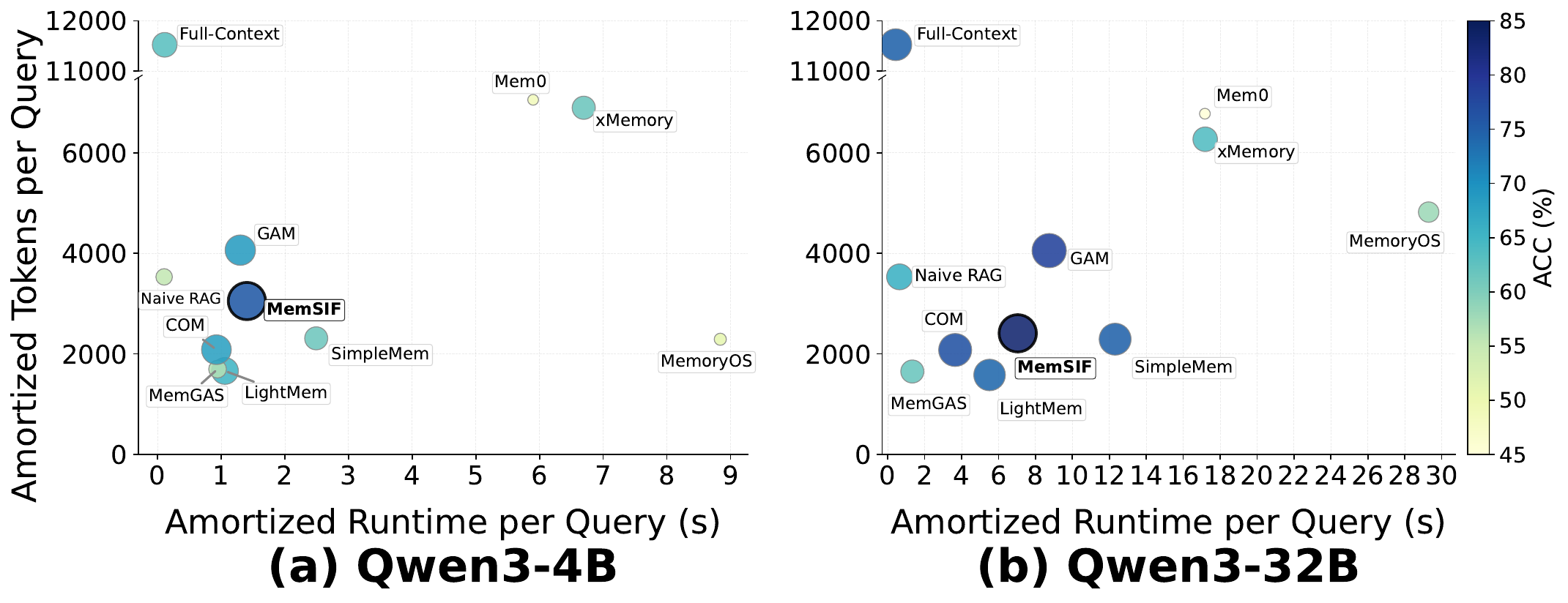}
\caption{Accuracy--cost trade-off on LoCoMo under Qwen3-4B and Qwen3-32B. 
 darker, larger bubbles indicate higher ACC.}
\label{fig:efficiency}
\end{figure}

\subsection{Efficiency Analysis}
\label{sec:efficiency}

We evaluate amortized tokens and runtime per query, covering the full pipeline
from memory construction to answer generation; detailed measurement scope is in
Appendix~\ref{app:efficiency_details}. Figure~\ref{fig:efficiency} reports the
accuracy--cost trade-off on LoCoMo. MemSIF achieves the highest ACC while
offering a favorable accuracy--cost trade-off under both Qwen3-4B and
Qwen3-32B. Under Qwen3-32B, it uses 40.6\% fewer tokens than GAM (2.41K
vs.\ 4.06K) while improving ACC from 77.21\% to 82.99\%; compared with
SimpleMem at a similar token budget, it improves ACC by 8.38\% and reduces
runtime by 42.8\% (7.04 vs.\ 12.31s). This efficiency comes from the
dual-track design: CoreFact memory stores compact reusable facts, while
persistent ActiveFact entries are reused across queries, reducing repeated
access to raw interactions and Structured Interaction Memory. Complete
numerical results are in Appendix~\ref{app:efficiency_details}.

\section{Conclusion}
\label{sec:conclusion}

We introduced \textsc{MemSIF}, a structured interaction-to-fact memory
framework that mitigates Temporal--Structural Misalignment and Delayed Utility
Manifestation through Structured Interaction Memory and Dual-Track Fact
Memory. Experiments on LoCoMo and LongMemEval-S across five backbone LLMs show
consistent Total ACC gains. Ablation and diagnostic results support the
complementary contributions of these two components, while efficiency analyses
show a favorable accuracy--cost trade-off. More broadly, LLM memory systems
should move beyond retrieval efficiency to explicitly model how interaction
experience evolves into reusable knowledge over time. \textsc{MemSIF} takes a
step in this direction. The NCE and LSHU diagnostic subsets provide a reusable framework for evaluating these misalignment patterns. Future work includes online CoreFact schema adaptation and extension to multi-agent settings.

\bibliography{aaai2027}
% Force appendix to start on a new page for clean PDF splitting
\clearpage
\appendix

\captionsetup{justification=raggedright, singlelinecheck=false}
\newcolumntype{Y}{>{\raggedright\arraybackslash}X}

\section{Empirical Analysis of Temporal--Structural Misalignment and Delayed Utility Manifestation}
\label{app:phenomena}

In brief, this section demonstrates that TSM and DUM are observable in
method-independent diagnostics. For TSM, NCE construction and ASV/SNTD analyses
show that chronological proximity is an incomplete proxy for topical or
event-level structure. For DUM, LSHU construction, LSHUR, and controlled
compaction diagnostics show that write-time salience often underestimates future
utility, affecting 43.71\% of queries with at least one critical evidence
utterance. The connection between these findings and MemSIF's architectural
responses is discussed in Section~\ref{app:a6}.

\subsection{Purpose and Connections to the Main Text}
\label{app:a1}

This appendix provides the empirical basis for the TSM and DUM claims
introduced in Section~\ref{sec:intro}. It defines the diagnostic measurements,
describes the construction of the Non-Contiguous Evidence (NCE) and
Low-Salience, High-Utility (LSHU) subsets used in the main text, and presents
the empirical diagnostics underlying Figure~\ref{fig:motivation}. The
diagnostic definitions and subset construction are independent of method
outputs, and method performance is referenced only to connect these fixed
subsets to Figure~\ref{fig:motivation}.

\subsection{Experimental Foundation}
\label{app:a2}

The analysis uses two long-horizon dialogue benchmarks with complementary
properties. LoCoMo provides densely annotated gold evidence over shared
multi-session conversations, making it suitable for both TSM and DUM analyses.
After excluding adversarial questions following the standard protocol, it
contains 1,540 evidence-grounded QA samples from 10 two-party conversations,
each spanning approximately 27 sessions and averaging 16.6K tokens.
LongMemEval-S provides a complementary large-context setting, where each sample
is paired with an independent interaction history averaging over 100K tokens.
After excluding 30 abstention questions, 470 evaluation samples remain. TSM is
analyzed on both datasets, while DUM is analyzed only on LoCoMo because it
requires gold evidence annotations.

For the TSM analysis, raw dialogue turns are consolidated into fixed-size
segments, with segment length $\ell$ varied over $\{1,2,4,8\}$ to test
robustness to analysis granularity. This segmentation is used only for empirical diagnosis
and is independent of MemSIF's semantic segmentation module, avoiding bias
toward the proposed method. All segments are encoded with Qwen3-Embedding-8B,
and cosine similarity between normalized embeddings defines the semantic space
used by ASV and SNTD@$k$.

For the DUM analysis, LoCoMo gold evidence utterances are evaluated along two
axes, write-time salience and query-time utility. Salience is scored without
revealing the future query, while utility is scored with the query but without
the gold answer. This design separates whether an utterance appears worth
storing when first observed from whether it later becomes useful for answering
a specific query.

For TSM, GPT-4o provides post-hoc validation that embedding-based measurements
correspond to genuine topical shifts or same-thread relations, using blind
judging and $N=3$ self-consistency majority voting. For DUM, salience and
utility scores are assigned by GPT-4o with a single call at temperature $0.1$,
where salience scoring is query-hidden and utility scoring uses the future
query but excludes the gold answer. Diagnostic detail-recoverability
and answer-correctness judgments use GPT-4o with blind evaluation,
deterministic input shuffling where applicable, and $N=3$ majority voting.
Additional decoding settings, prompt templates, random seeds, and computational
costs are reported in Section~\ref{app:a5}.

\begin{figure*}[t]
\centering
\includegraphics[width=\textwidth]{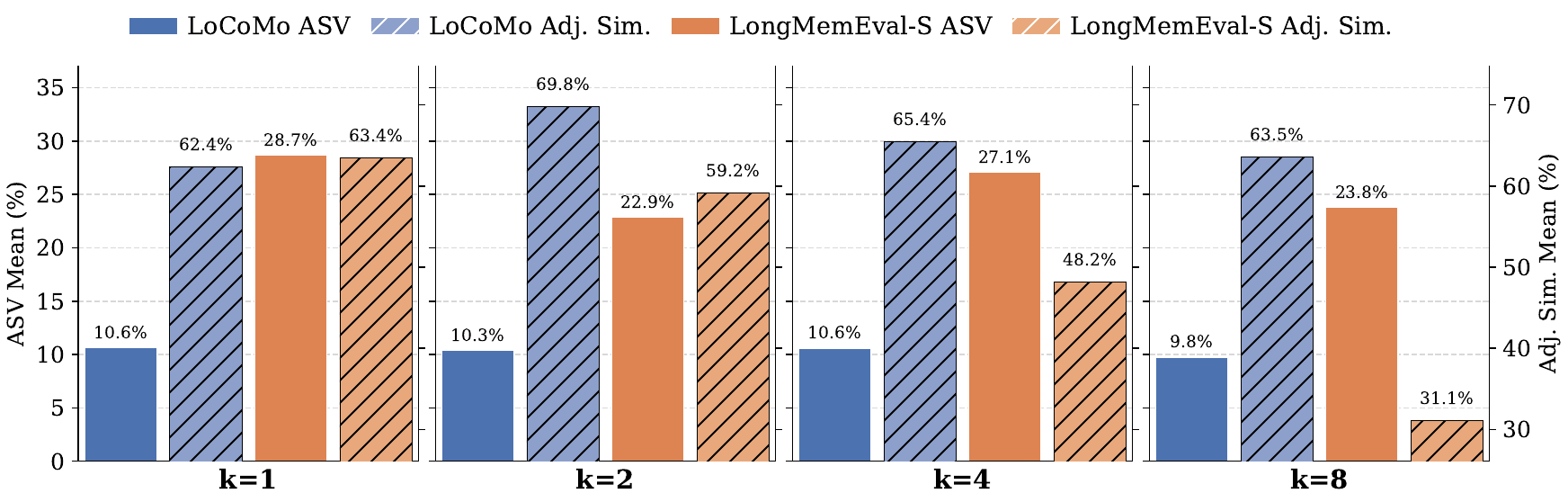}
\caption{Adjacent Semantic Volatility (ASV) and adjacent similarity across segment lengths on LoCoMo and LongMemEval-S.}
\label{fig:asv_chunk_sensitivity}
\end{figure*}

\subsection{Diagnostic Subset Construction}
\label{app:diagnostic_subsets}

We construct two diagnostic subsets from the 1,540 non-adversarial LoCoMo
questions to evaluate Temporal--Structural Misalignment and Delayed Utility
Manifestation more directly. Both subsets are constructed independently of
MemSIF and baseline outputs, and the same fixed question IDs are used in
Figures~\ref{fig:motivation} and~\ref{fig:ablation_linechart}.
NCE operationalizes the cross-time evidence-fragmentation aspect of TSM, while
LSHU operationalizes DUM at the query level by selecting questions whose
critical evidence was not salient when first observed.

\paragraph{Non-Contiguous Evidence.}
For each question $q$, let $\mathcal{E}_q$ denote its gold evidence utterances
and let $\mathrm{pos}(e)$ denote the global message position of evidence
utterance $e$ in an interaction history of length $L_q$. We define Temporal
Evidence Dispersion (TED) as
\begin{equation}
\mathrm{TED}(q)=
\frac{
\max_{e\in\mathcal{E}_q}\mathrm{pos}(e)
-
\min_{e\in\mathcal{E}_q}\mathrm{pos}(e)
}{
L_q-1
}.
\end{equation}
TED lies in $[0,1]$ and measures the normalized span covered by the evidence,
allowing comparison across conversations of different lengths. We first retain
the 409 questions containing at least two gold evidence utterances. Their TED
distribution is concentrated near very small spans and large spans, with a
relatively sparse transition region, with 156 questions falling in $[0,0.1)$, 67 in
$[0.1,0.3)$, and 186 in $[0.3,1]$. We therefore use
$\mathrm{TED}(q)\geq0.3$, a threshold located at the valley between the two
modes of the empirical TED distribution, yielding 186 questions. The selected
questions have mean TED $0.581$ and mean evidence span $352.6$ message
positions.

\paragraph{Low-Salience, High-Utility.}
For each gold evidence utterance $e$, GPT-4o assigns a write-time salience
score $S(e)\in\{1,2,3\}$ using only $e$ and its two preceding utterances.
Because salience is intended to represent write-time value, the query is hidden
during this step and the score is assigned independently of any future query.
For each evidence--query pair $(e,q)$, GPT-4o separately assigns a
query-time utility score $U(e,q)\in\{1,2,3\}$ using the same local context
together with $q$. The gold answer is excluded from both scoring procedures.
All scores are obtained with a single call at temperature $0.1$.

A score of $S(e)\leq2$ indicates that the utterance has no clear archival value
when first observed, whereas $U(e,q)=3$ indicates that it is critical for
answering $q$. We define the strict Low-Salience, High-Utility (LSHU) subset as
\begin{equation}
\mathcal{D}_{\mathrm{LSHU}}
=
\left\{
q:
\exists e\in\mathcal{E}_q,\,
S(e)\leq2
\land
U(e,q)=3
\right\}.
\end{equation}
Across 2,229 evidence--query pairs, 1,035 receive $U(e,q)=3$, of which 430
also satisfy $S(e)\leq2$, giving a pair-level LSHU rate of $41.55\%$.
At the question level, 970 questions contain at least one critical evidence
utterance, and 424 contain at least one strict LSHU utterance. The resulting
query-level rate is $43.71\%$ among high-utility questions and $27.53\%$ among
all 1,540 evaluation questions.

Table~\ref{tab:diagnostic_subset_stats} reports the category distributions of
the resulting subsets. NCE is dominated by M-hop questions, consistent with
their reliance on evidence distributed across interaction units. LSHU is
more concentrated in S-hop and Temp questions, suggesting that delayed utility
is not restricted to multi-hop evidence composition.

\begin{table}[t]
\centering
\caption{Question category distributions of the diagnostic subsets.}
\label{tab:diagnostic_subset_stats}
\resizebox{\columnwidth}{!}{%
\small\setlength{\tabcolsep}{5pt}
\begin{tabular}{lccccc}
\toprule
\textbf{Subset} & \textbf{S-hop} & \textbf{M-hop} &
\textbf{Temp} & \textbf{Kno} & \textbf{Total} \\
\midrule
NCE
& 0 (0.0\%)
& 152 (81.7\%)
& 12 (6.5\%)
& 22 (11.8\%)
& 186 \\
LSHU
& 271 (63.9\%)
& 44 (10.4\%)
& 109 (25.7\%)
& 0 (0.0\%)
& 424 \\
\bottomrule
\end{tabular}%
}
\end{table}

The complete question-ID lists are provided in the supplementary materials,
enabling exact reproduction of both diagnostic evaluations.
These fixed subsets are used for the diagnostic comparisons in
Figure~\ref{fig:motivation}.

\subsection{Temporal--Structural Misalignment}
\label{app:a3}

In long-horizon dialogues, chronological order does not consistently indicate
topical continuity, since adjacent turns may mix unrelated topics while segments
sharing the same event or task can appear at widely separated positions. This
section examines whether temporal order alone suffices to capture the topical
organization of such dialogues.

The analysis follows a construct-proxy-validation framework. The core
construct, topical relatedness, captures whether segments share an event or
task thread. Adjacent Semantic Volatility (ASV) and Semantic-Neighbor Temporal
Distance (SNTD@$k$) serve as embedding-based operational proxies; LLM judgments
independently validate that these proxies track the intended construct,
distinguishing meaningful topical shifts from surface-level variation.

Two complementary claims are examined. Claim~1.1
(Section~\ref{app:a3_1}) tests whether temporal adjacency guarantees topical
continuity, using ASV and semantic-shift validation. Claim~1.2
(Section~\ref{app:a3_2}) tests whether topical relatedness is confined to
local proximity, using SNTD@$k$ and same-thread validation.

The LoCoMo and LongMemEval-S corpora introduced in Section~\ref{app:a2} are
used, with segment embeddings computed via Qwen3-Embedding-8B. LLM-based
validation employs GPT-4o with blind judging and $N=3$ majority voting, as
detailed in Section~\ref{app:a2}.

\subsubsection{Claim 1.1: Temporal Adjacency Does Not Guarantee Semantic Continuity}
\label{app:a3_1}

Claim~1.1 tests whether temporal adjacency consistently indicates topical
continuity. If temporal adjacency were a reliable proxy for topical continuity,
adjacent segment similarities would exhibit infrequent sharp changes. Adjacent semantic similarity variation thus
serves as a measurable proxy for local topical discontinuity.

This is quantified through Adjacent Semantic Volatility (ASV). Given a segment sequence $s_1, \ldots, s_m$ under a fixed segment length, with normalized embeddings $e_1, \ldots, e_m$, the cosine similarity between adjacent segments is:
\begin{equation}
a_i = \cos(e_i, e_{i+1}), \quad i = 1, \ldots, m-1.
\end{equation}
Let $A = \{a_1, \ldots, a_{m-1}\}$ denote the adjacent similarity sequence. ASV measures the mean absolute fluctuation:
\begin{equation}
\mathrm{ASV} = \frac{1}{m-2} \sum_{i=1}^{m-2} |a_{i+1} - a_i|.
\end{equation}
Higher ASV indicates greater fluctuation in adjacent similarity, reflecting weaker local topical continuity.

ASV is computed per conversation using the segment embeddings from Section~\ref{app:a2} and aggregated at the dataset level, repeating under segment lengths $\ell \in \{1, 2, 4, 8\}$ for robustness. Figure~\ref{fig:asv_chunk_sensitivity} reports the results.

Across all segment lengths, both datasets display non-trivial volatility. If
temporal adjacency consistently indicated topical continuity, adjacent
similarities would show limited local fluctuation. The observed ASV, stable
across granularities, thus indicates that adjacency is not a consistent proxy
for topical continuity. LongMemEval-S's consistently higher ASV further
suggests that denser interaction histories weaken this consistency.

To verify that ASV captures authentic topical discontinuity, LLM semantic-shift
validation is conducted. Adjacent pairs under $\ell=2$ are bucketed by a local
pattern score
\begin{equation}
r_i = z(1-a_i) + z(\Delta_i),
\quad
\Delta_i =
\max\{|a_i-a_{i-1}|, |a_{i+1}-a_i|\},
\end{equation}
where unavailable boundary terms are omitted and $z(\cdot)$ denotes
within-conversation standardization. The high-volatility/low-similarity bucket
is sampled from the top tercile of $r_i$, the low-volatility/high-similarity
bucket from the bottom tercile, and the mid bucket from the middle tercile.
This contrasts stable local continuity against sharp local semantic drops. For
each bucket, 100 pairs are judged by GPT-4o with $N=3$ majority voting as
described in Section~\ref{app:a2}.
Table~\ref{tab:asv_llm_validation} reports the results.

\begin{table}[t]
\centering
\caption{LLM semantic-shift validation by volatility bucket.}
\label{tab:asv_llm_validation}
\resizebox{\columnwidth}{!}{%
\small\setlength{\tabcolsep}{4pt}
\begin{tabular}{lcc}
\toprule
\textbf{Dataset} & \textbf{Bucket} & \textbf{Semantic Shift Rate} \\
\midrule
LoCoMo & High-volatility / low-similarity & 23.00\% \\
LoCoMo & Mid & 3.00\% \\
LoCoMo & Low-volatility / high-similarity & 3.00\% \\
LongMemEval-S & High-volatility / low-similarity & 47.00\% \\
LongMemEval-S & Mid & 20.00\% \\
LongMemEval-S & Low-volatility / high-similarity & 12.00\% \\
\bottomrule
\end{tabular}%
}
\end{table}

LLM validation supports the use of ASV as a proxy for topical discontinuity:
semantic shift rates are notably higher in high-volatility than
low-volatility buckets for both datasets, with 23\% versus 3\% for LoCoMo and
47\% versus 12\% for LongMemEval-S. The higher shift rates in the
high-volatility buckets suggest that ASV captures meaningful topical
discontinuities rather than only embedding noise. The embedding-based
volatility metric thus provides evidence that memory systems should model
topical structure rather than rely solely on chronological adjacency.

These results indicate that temporal adjacency does not consistently guarantee topical continuity. Temporal order alone is insufficient for capturing topical structure in long-horizon memory.

\subsubsection{Claim 1.2: Temporal Dispersion of Semantic Neighbors}
\label{app:a3_2}

This section examines whether topically related segments are confined to local
proximity. If local temporal windows sufficed, semantic neighbors should
cluster nearby; however, information about the same event, task, or character
can recur across sessions at widely separated positions. The analysis tests
whether topical relatedness is fully captured by local temporal proximity:
SNTD@$k$ serves as the operational proxy via embedding similarity, and LLM
same-thread validation evaluates whether temporally distant semantic neighbors
reflect authentic topical threads.

\begin{figure*}[t]
\centering
\includegraphics[width=\textwidth]{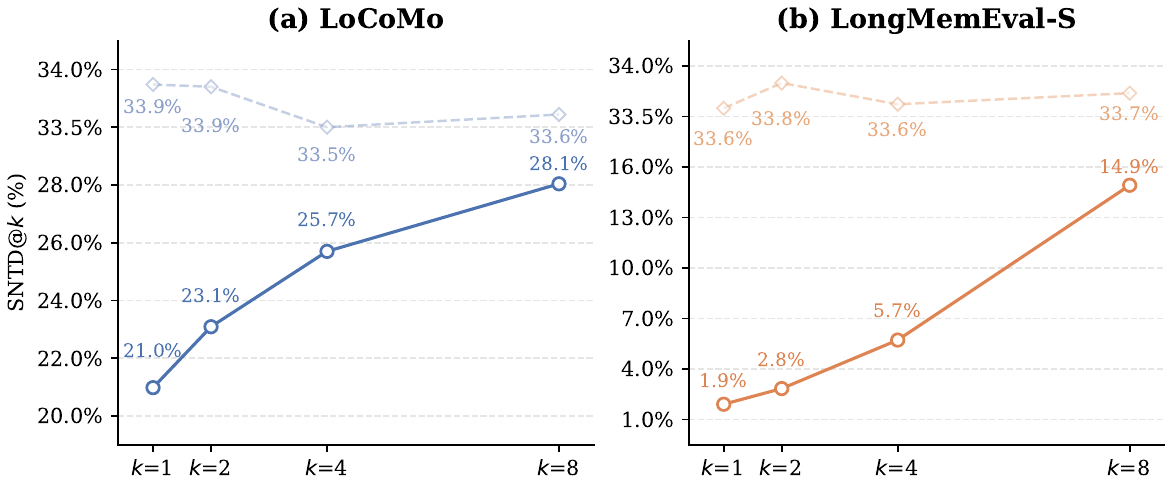}
\caption{Semantic-Neighbor Temporal Distance (SNTD@$k$) compared with random baseline on LoCoMo and LongMemEval-S.}
\label{fig:sntd_trend}
\end{figure*}

This is quantified through Semantic-Neighbor Temporal Distance at rank $k$
(SNTD@$k$). Given a segment sequence $s_1, \ldots, s_m$ under a fixed segment
length $\ell=2$, with normalized embeddings $e_1, \ldots, e_m$, the top-$k$
semantic neighbors $\mathcal{N}_k(i)$ of segment $i$ are retrieved within the
same conversation by cosine similarity, excluding $i$ itself. The normalized
temporal distance between segment $i$ and neighbor $j$ is:
\begin{equation}
d(i, j) = \frac{|i - j|}{m - 1}.
\end{equation}
SNTD@$k$ is the mean normalized temporal distance across all top-$k$ semantic neighbors:
\begin{equation}
\mathrm{SNTD}@k = \frac{1}{m} \sum_{i=1}^{m} \frac{1}{k} \sum_{j \in \mathcal{N}_k(i)} d(i, j).
\end{equation}
Higher SNTD@$k$ indicates wider temporal dispersion of semantic neighbors. A
random baseline, Random SNTD@$k$, is constructed by drawing $k$ non-self
segments uniformly at random. If SNTD@$k$ is substantially lower than the
random baseline, semantic neighbors are locally concentrated; if SNTD@$k$
increases with $k$, broader semantic retrieval is reaching more temporally
distant neighbors. Following the construct-proxy-validation framework,
SNTD@$k$ is computed per conversation from the embeddings of Section~\ref{app:a2}, under neighbor ranks $k \in \{1, 2, 4, 8\}$, with the random baseline computed identically. Retrieval is restricted to the same conversation. Figure~\ref{fig:sntd_trend} reports results.

Both datasets show that semantic neighbors are not randomly scattered. LoCoMo
exhibits local concentration, with SNTD@1 at 20.98\% versus the random baseline
of 33.87\%. However, SNTD@$k$ rises from 20.98\% to 28.11\% as $k$ increases
from 1 to 8, approaching the random baseline. LongMemEval-S shows stronger
local concentration, with SNTD@1 at 1.91\%, yet follows the same monotonic
trend from 1.91\% to 14.91\%. These results indicate that the closest semantic
neighbors are generally temporally concentrated, but broader semantic retrieval
progressively covers more distant segments. This increase is interpreted
descriptively, since including lower-ranked neighbors naturally expands
temporal coverage; by itself, it does not establish that the distant neighbors
belong to the same topical thread. We therefore use LLM same-thread judgments
to evaluate whether temporally distant semantic neighbors represent genuine
cross-time associations.

To verify whether distant semantic neighbors reflect authentic topical
threads, LLM same-thread validation is conducted following the same protocol as
Section~\ref{app:a3_1}. Using segment length $\ell=2$, semantic-neighbor pairs
are divided into near/mid/far temporal-distance buckets; GPT-4o judges
same-thread membership with $N=3$ voting. Table~\ref{tab:sntd_llm_validation}
reports the results.

\begin{table}[t]
\centering
\caption{LLM same-thread validation by temporal-distance bucket.}
\label{tab:sntd_llm_validation}
\resizebox{\columnwidth}{!}{%
\small\setlength{\tabcolsep}{4pt}
\begin{tabular}{lcc}
\toprule
\textbf{Dataset} & \textbf{Distance Bucket} & \textbf{Same-thread Rate} \\
\midrule
LoCoMo & Near & 82.00\% \\
LoCoMo & Mid & 74.00\% \\
LoCoMo & Far & 70.00\% \\
LongMemEval-S & Near & 59.00\% \\
LongMemEval-S & Mid & 37.00\% \\
LongMemEval-S & Far & 5.00\% \\
\bottomrule
\end{tabular}%
}
\end{table}

In LoCoMo, same-thread rates remain high across all buckets, with near at
82.00\%, mid at 74.00\%, and far at 70.00\%, indicating robust cross-temporal
topical continuity. In LongMemEval-S, rates decline sharply with distance, from
59.00\% at near range to 5.00\% at far range, indicating stronger local topical
concentration. The two datasets therefore expose different degrees of temporal
dispersion, where LoCoMo contains robust cross-time topical continuity while
LongMemEval-S is more locally concentrated. In both cases, however, temporal
position alone is an incomplete description of topical structure, because
semantic relatedness must be modeled separately from raw chronology.

These results indicate that topical relatedness is not fully captured by local
temporal proximity. While both datasets exhibit local concentration, semantic
structure cannot be reduced to narrow chronological windows. Local proximity is
useful but insufficient for representing the topical structure of long-horizon
dialogues.

Taken together, Section~\ref{app:a3} provides evidence for
Temporal--Structural Misalignment from two complementary angles. ASV analysis
indicates non-trivial adjacent volatility, with LLM validation confirming that
high-volatility pairs are more likely to correspond to topical discontinuity,
indicating that temporal adjacency does not consistently guarantee topical
continuity. SNTD@$k$ analysis shows that semantic neighbors are locally
concentrated yet can span progressively wider temporal distances, with LLM
validation demonstrating dataset-dependent degrees of cross-time topical
continuity. LoCoMo provides strong evidence for recurring cross-time event
threads, while LongMemEval-S shows a more local structure; together they
indicate that chronological order alone does not describe topical organization.
Section~\ref{app:a6} discusses how these findings motivate MemSIF's
architectural responses.

\subsection{Delayed Utility Manifestation}
\label{app:a4}

In long-horizon dialogues, the utility of an interaction is often not apparent
when it first occurs, since segments that appear as low-salience background details
at write time may later become critical evidence for answering future queries.
Write-time salience is therefore an imperfect guide to future query-time
utility. This misalignment has direct consequences for memory systems that
commit to retention or compaction decisions before future information needs are
known.

Following the construct-proxy-validation framework introduced in
Section~\ref{app:a3}, the core construct here is the predictability of future
utility---that is, whether the eventual value of information can be
consistently judged at write time. Low-Salience High-Utility Rate (LSHUR),
Compaction Detail Loss (CDL), and Query Answerability Drop (QAD) serve as
operational proxies computed from LLM-assisted judgments under the consistency
controls of Section~\ref{app:a2}. CDL-QAD correlation provides an internal
consistency check.

Two complementary claims are examined. Claim~2.1
(Section~\ref{app:a4_1}) tests whether write-time salience aligns with future
utility, using LSHUR as the proxy. Claim~2.2 (Section~\ref{app:a4_2}) tests
whether premature compaction reduces detail recoverability and answerability,
using CDL and QAD as proxies with CDL-QAD correlation for validation.

The LoCoMo corpus is used exclusively, as gold evidence annotations are
required for salience and utility scoring. Salience and utility scores are
assigned by GPT-4o with a single call at temperature $0.1$, following the
query-hidden and gold-answer-hidden protocol described in
Section~\ref{app:a2}. Diagnostic detail-recoverability and answer-correctness
judgments use GPT-4o with blind evaluation and $N=3$ majority voting.

\subsubsection{Claim 2.1: Write-Time Salience Is Misaligned with Future Utility}
\label{app:a4_1}

Claim~2.1 tests whether write-time salience consistently predicts future
query-time utility. If write-time salience sufficed, future high-utility
evidence would consistently appear high-salience when first observed; a
systematic mismatch would indicate that write-time value assessment is
inherently incomplete.

This is quantified through Low-Salience High-Utility Rate (LSHUR). For each
evidence-query pair $(e, q)$, write-time salience score
$S(e) \in \{1, 2, 3\}$ and query-time utility score
$U(e, q) \in \{1, 2, 3\}$ are independently assigned. $S = 3$ indicates the
utterance merits immediate archival as a durable fact; $S \le 2$ indicates no
clear immediate archival value. $U = 3$ indicates critical/direct evidence for
answering the future query. At the pair level, LSHUR is the proportion of
future high-utility pairs whose write-time salience falls below threshold
$\tau_s$:
\begin{equation}
\mathrm{LSHUR}_{\mathrm{pair}} = \frac{|\{(e,q): S(e) \le \tau_s \land U(e,q) \ge \tau_u\}|}{|\{(e,q): U(e,q) \ge \tau_u\}|}.
\end{equation}
At the query level, LSHUR counts queries with at least one low-salience high-utility evidence:
\begin{equation}
\mathrm{LSHUR}_{\mathrm{query}} = \frac{|\{q: \exists e,\ S(e) \le \tau_s \land U(e,q) \ge \tau_u\}|}{|\{q: \exists e,\ U(e,q) \ge \tau_u\}|}.
\end{equation}
The low-salience threshold is fixed at $\tau_s = 2$ (i.e., $S(e) \le 2$), and
the utility threshold is fixed at $\tau_u = 3$, admitting only evidence judged
critical for answering the future query.

Write-time salience is scored with the LLM seeing only the target utterance and
its two preceding utterances, with no future query, gold answer, or downstream
task provided, ensuring query-agnostic scoring. Query-time utility scoring adds
the future query but excludes the gold answer. Salience and utility scores are
assigned by GPT-4o with a single call at temperature $0.1$, following
Section~\ref{app:a2}. Table~\ref{tab:lshur_overall} reports overall LSHUR.

\begin{table}[t]
\centering
\caption{Overall Low-Salience High-Utility Rate (LSHUR).}
\label{tab:lshur_overall}
\resizebox{\columnwidth}{!}{%
\small\setlength{\tabcolsep}{4pt}
\begin{tabular}{lcc}
\toprule
\textbf{Metric} & \textbf{Numerator / Denominator} & \textbf{Rate} \\
\midrule
LSHUR-pair & 430 / 1,035 & 41.55\% \\
LSHUR-query & 424 / 970 & 43.71\% \\
\bottomrule
\end{tabular}%
}
\end{table}

Results reveal a substantial salience-utility misalignment across the full
LoCoMo evaluation set. At the pair level, 41.55\% of future-critical
evidence--query pairs are low-salience at write time. At the query level,
43.71\% of queries with at least one critical evidence utterance contain at
least one such low-salience critical utterance. If write-time salience were a
reliable proxy for future utility, the proportion of low-salience high-utility
evidence would be expected to be substantially lower. The observed rates
instead indicate that value assessment at write time is often incomplete,
suggesting that memory architectures should avoid relying solely on write-time
retention decisions.

The category distribution of the resulting LSHU questions is provided in
Table~\ref{tab:diagnostic_subset_stats}. The subset is concentrated in S-hop
and Temp questions, indicating that delayed utility extends beyond
multi-hop evidence composition. Because salience scoring is query-hidden,
utility scoring excludes the gold answer, and neither score uses method
outputs, the subset construction does not depend on gold-answer leakage or
method performance.

These results indicate that write-time salience is an imperfect proxy for
future utility. A non-trivial fraction of future-critical evidence is misjudged
at write time, affecting 43.71\% of queries that contain at least one critical
evidence utterance.

\subsubsection{Claim 2.2: Premature Compaction Reduces Detail Recoverability and Answerability}
\label{app:a4_2}

This subsection examines whether the salience-utility misalignment established
in Claim~2.1 produces measurable consequences. If a system performs early
compaction based on write-time salience, before future queries reveal actual
needs, query-relevant details may be lost, degrading downstream answerability.

Two metrics are defined. Compaction Detail Loss (CDL) measures the proportion
of query-critical gold facts unrecoverable from compacted memory. For query
$q$ with gold fact set $\mathcal{G}_q$ and compacted memory $M_c$, an LLM judge
classifies each fact as fully recoverable, partially recoverable, or not
recoverable. The Detail Recovery Rate (DRR) yields
$\mathrm{CDL}_c(q) = 1 - \mathrm{DRR}_c(q).$ Query Answerability Drop (QAD)
measures the relative degradation in answer correctness when using compacted
memory versus raw gold evidence. For each query, answers are generated from raw
evidence and from compacted memory $M_c$; an LLM judge performs blind
correctness evaluation. CDL is computed at the fact level and then aggregated,
whereas QAD is computed at the query level.

Both metrics report strict and lenient variants. Under the strict setting, partially correct/recoverable counts as incorrect/unrecoverable. Under the lenient setting, it counts as 0.5. QAD is then $\mathrm{QAD}_c = \mathrm{Acc}_{\mathrm{raw}} - \mathrm{Acc}_c.$

CDL and QAD are evaluated on 100 queries sampled from the strict LSHU subset
using question-category stratification.
Premature compaction is simulated at the query level via salience-ranked
truncation. For each query $q$, its gold evidence utterances are ranked by
their query-hidden write-time salience score $S(e)$, and only the top-$c$
utterances are retained as compacted memory $M_c$ for
$c \in \{2, 4, 8\}$. This models query-agnostic compaction without knowledge of
future information needs. This setup is not intended as a deployable memory
baseline; it is a controlled diagnostic that isolates the effect of
salience-ranked retention from retrieval errors. Even when the candidate pool
is restricted to gold evidence, write-time salience alone fails to preserve all
query-critical details. Query-critical gold facts are drawn from LoCoMo gold
evidence annotations. Answer generation uses GPT-4o-mini; all recoverability
and correctness judgments use GPT-4o under the protocol described in
Section~\ref{app:a2}.

\paragraph{Detail Recoverability (CDL).}
Table~\ref{tab:cdl} reports CDL across memory sources.

\begin{table*}[t]
\centering
\caption{Compaction Detail Loss (CDL) across memory sources.}
\label{tab:cdl}
\resizebox{\textwidth}{!}{%
\small\setlength{\tabcolsep}{4pt}
\begin{tabular}{lccccccc}
\toprule
\textbf{Memory Source} & \textbf{Fully Recov.} & \textbf{Partially Recov.} & \textbf{Not Recov.} & \textbf{DRR Strict} & \textbf{CDL Strict} & \textbf{DRR Lenient} & \textbf{CDL Lenient} \\
\midrule
Raw evidence & 120 & 15 & 11 & 82.19\% & 17.81\% & 87.33\% & 12.67\% \\
$c = 2$ & 64 & 4 & 78 & 43.84\% & 56.16\% & 45.21\% & 54.79\% \\
$c = 4$ & 73 & 3 & 70 & 50.00\% & 50.00\% & 51.03\% & 48.97\% \\
$c = 8$ & 74 & 6 & 66 & 50.68\% & 49.32\% & 52.74\% & 47.26\% \\
\bottomrule
\end{tabular}%
}
\end{table*}

Compacted memory reduces detail recoverability. The raw evidence baseline
serves as a diagnostic upper-bound source, since recoverability is still judged
from natural-language evidence under the same blind LLM protocol. It achieves a strict CDL of 17.81\%, while all compaction
settings raise strict CDL to 49\%--56\%. Even under lenient criteria, CDL
remains above 47\%. When compaction decisions are made at write time without
knowledge of future queries, nearly half of the gold-standard evidence becomes
unrecoverable. The gap between the raw baseline and all compaction settings
indicates that write-time salience, as the sole retention criterion, is prone
to discarding query-relevant details.

\paragraph{Answerability Degradation (QAD).}
Table~\ref{tab:qad} reports QAD across memory sources.

\begin{table*}[t]
\centering
\caption{Query Answerability Drop (QAD) across memory sources.}
\label{tab:qad}
\resizebox{\textwidth}{!}{%
\small\setlength{\tabcolsep}{4pt}
\begin{tabular}{lccccccc}
\toprule
\textbf{Memory Source} & \textbf{Correct} & \textbf{Partially Correct} & \textbf{Incorrect} & \textbf{Acc Strict} & \textbf{Acc Lenient} & \textbf{QAD Strict} & \textbf{QAD Lenient} \\
\midrule
Raw gold evidence & 56 & 21 & 23 & 56.00\% & 66.50\% & --- & --- \\
$c = 2$ & 31 & 25 & 44 & 31.00\% & 43.50\% & 25.00\% & 23.00\% \\
$c = 4$ & 31 & 35 & 34 & 31.00\% & 48.50\% & 25.00\% & 18.00\% \\
$c = 8$ & 40 & 29 & 31 & 40.00\% & 54.50\% & 16.00\% & 12.00\% \\
\bottomrule
\end{tabular}%
}
\end{table*}

The answerability degrades measurably under compaction. Raw gold evidence achieves strict accuracy of 56.00\%; compacted memory accuracy drops to 31\%--40\% across $c$ settings, yielding QAD strict of 16\%--25\%. Lenient QAD shows consistent attenuation of 12\%--23\%. This degradation indicates that the detail loss measured by CDL translates into tangible downstream harm: answers generated from compacted memory are measurably less accurate, and the effect persists even under lenient evaluation criteria that credit partial correctness.

\begin{figure*}[t]
\centering
\includegraphics[width=\textwidth]{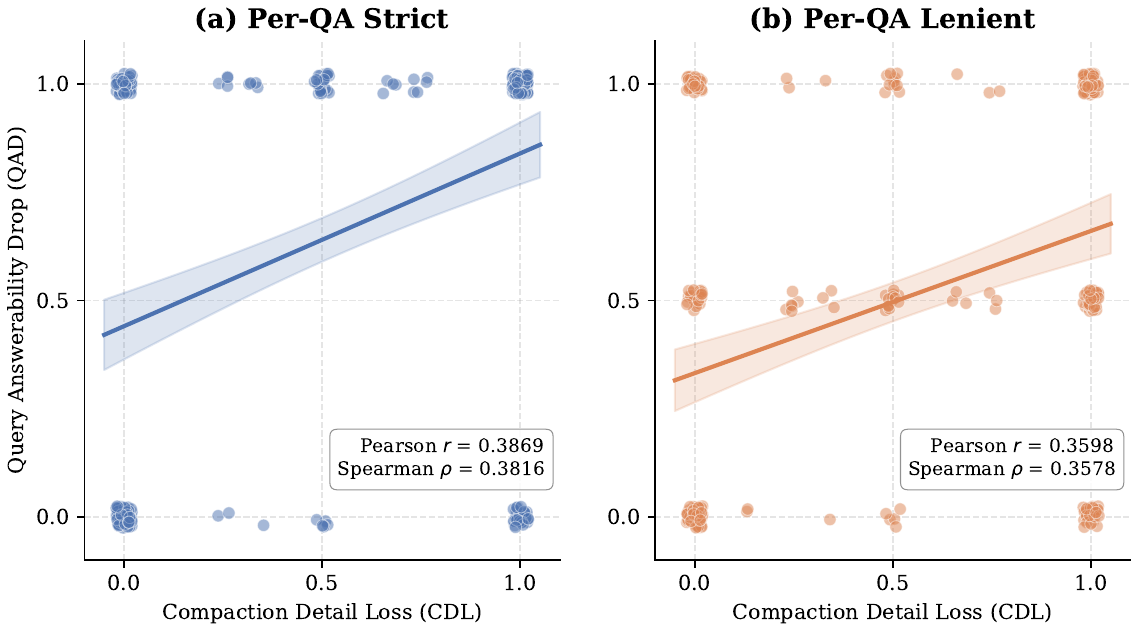}
\caption{Compaction Detail Loss--Query Answerability Drop (CDL--QAD) correlation at per-QA level (300 observations = 100 queries $\times$ 3 compaction settings $c \in \{2,4,8\}$). QAD values are discrete (strict: $\{0.0, 1.0\}$, lenient: $\{0.0, 0.5, 1.0\}$); a small vertical jitter ($\pm$0.025) is applied to reveal point density. All reported statistics (Pearson $r$, Spearman $\rho$) are computed on the original unjittered data.}
\label{fig:cdl_qad_scatter}
\end{figure*}

\paragraph{CDL--QAD Correlation.}
To verify that memory-level loss and task-level degradation are internally consistent, CDL-QAD correlation is computed at both aggregate and per-QA levels. Figure~\ref{fig:cdl_qad_scatter} reports the per-QA results.

At the aggregate level, Pearson $r$ is positive under both settings (strict:
$r = 0.58$; lenient: $r = 0.94$). At the per-QA level spanning 300 observations
from 100 queries across three compaction settings, correlations are moderate
but consistently positive (strict: $r = 0.39$; lenient: $r = 0.36$). These
correlations show that memory-level detail loss and task-level degradation
co-vary consistently, while residual per-instance variability remains.

These results indicate that premature compaction based on write-time salience
substantially reduces detail recoverability, with strict DRR dropping from
82.19\% under raw evidence to 43.84\%--50.68\% under salience-ranked
compaction, and degrades answerability by 16\%--25\%. The positive CDL-QAD
correlation indicates that memory-level loss is associated with task-level
degradation.

Taken together, Section~\ref{app:a4} provides evidence for Delayed Utility
Manifestation from two complementary angles. LSHUR analysis indicates that
write-time salience underestimates future query-time utility, with the gap
affecting 43.71\% of queries that contain at least one critical evidence
utterance. CDL and QAD analyses indicate that premature compaction based on
write-time salience produces measurable degradation in both detail
recoverability and answerability, with positive CDL-QAD correlation confirming
that these two dimensions of loss are internally consistent. These results
demonstrate that write-time salience is an imperfect proxy for future utility,
since information value often emerges only when future queries reveal specific
needs. Section~\ref{app:a6} discusses how these findings motivate MemSIF's
Dual-Track Fact Memory.

\subsection{Reproducibility and Implementation Details}
\label{app:a5}

Segment embeddings are computed with Qwen3-Embedding-8B. GPT-4o is used for
DUM salience and utility scoring as well as judgment-oriented validation.
Task-specific prompts, query-hidden salience scoring, gold-answer-hidden
utility scoring, and blind evaluation are used to reduce direct leakage and
method-output bias.

\paragraph{Decoding and Consistency Controls.} Salience and utility scores are
assigned with a single call at temperature $0.1$. Salience scoring is
query-hidden: the model sees only the target utterance and its two preceding
utterances. Utility scoring uses the future query but excludes the gold answer.
Judgment-oriented steps, including TSM semantic-shift validation, same-thread
validation, detail recoverability, and answer correctness, use temperature
$0.0$ with $N=3$ self-consistency majority voting. Correctness and
recoverability judgments use blind evaluation with deterministically shuffled
input order where applicable.

\paragraph{Computational Cost.} Embedding computation processes the full
LoCoMo corpus under each segment-length configuration; its cost is
deterministic and substantially lower than the LLM-based annotation and
judgment steps. TSM LLM validation involves 600 pairwise judgments (2 datasets
$\times$ 3 buckets $\times$ 100 pairs) at $\sim$200 tokens per judgment. DUM
salience and utility annotation includes utility scoring for 2,229
evidence--query pairs plus salience scoring for the corresponding evidence
utterances, for approximately 3,000 LLM calls in total. Answer generation for
QAD evaluation (100 queries $\times$ 4 memory sources = 400 generations)
consumes approximately 200K output tokens.

\paragraph{Reproducibility.} All sampling and bucket-construction operations,
including the 100-query stratified sample for CDL/QAD evaluation, use fixed
seed 42. Fixed-length segmentation, embedding computation, and statistical
aggregation are deterministic given the same inputs. LLM-based steps are
reproducible given the same model versions, prompts, decoding settings, and
released intermediate annotations. Full prompt templates and generated
annotation files are provided in the supplementary materials.

\subsection{From Empirical Findings to Architecture Design}
\label{app:a6}

The empirical findings of Sections~\ref{app:a3} and~\ref{app:a4} clarify why
MemSIF separates interaction organization from fact construction. TSM shows that
chronological adjacency alone cannot represent both local topic boundaries and
cross-time event continuity. This motivates Structured Interaction Memory, where
Topical Segments preserve locally coherent interaction units while Event
Trajectories reconnect non-contiguous units that belong to the same
evolving event or task.

DUM shows that write-time salience is an incomplete predictor of future utility
and that salience-based compaction can reduce both detail recoverability and
answerability. This motivates Dual-Track Fact Memory. CoreFact memory consolidates
schema-guided information whose reusable value can be identified at write time, while uncertain details
remain recoverable from Structured Interaction Memory. ActiveFact memory then
supports query-driven formation when later queries reveal the utility of those
details.

Together, these two designs address different points in the memory activation
and consolidation process. Structured Interaction Memory changes how interaction histories are organized,
whereas Dual-Track Fact Memory changes when uncertain information is activated
and consolidated. Queries reveal which recoverable details matter, and recurring
source support with recurring query demand can turn such details into
persistent ActiveFact entries.

\begin{table*}[t]
\centering
\caption{Main results on LoCoMo with Qwen3-8B and Llama-3.1-8B-Instruct. \textbf{Bold} indicates the best result and \uline{underlined} the second-best for each column.}
\label{tab:locomo_supplementary}
\resizebox{\textwidth}{!}{%
\small\setlength{\tabcolsep}{4pt}
\begin{tabular}{l|ccccc|ccccc}
\toprule
\multirow{2}{*}{Method}
  & \multicolumn{5}{c|}{Qwen3-8B}
  & \multicolumn{5}{c}{Llama-3.1-8B-Instruct} \\
\cmidrule(lr){2-6} \cmidrule(lr){7-11}
 & S-hop & M-hop & Temp & Kno & Total
 & S-hop & M-hop & Temp & Kno & Total \\
\midrule
Full-Context & \uline{86.49} & 58.16 & 44.37 & 33.33 & 69.21 & 77.75 & 47.66 & 50.12 & 58.54 & 65.28 \\
Naive RAG & 76.42 & 37.21 & 45.99 & 27.33 & 59.84 & 69.80 & 38.63 & 46.88 & 42.55 & 57.62 \\
Mem0 & 67.66 & 50.20 & 37.24 & 42.02 & 56.52 & 52.23 & 52.78 & 37.80 & 47.94 & 49.06 \\
MemoryOS & 70.84 & 51.65 & 32.88 & 46.52 & 57.90 & 56.41 & \uline{57.95} & 45.46 & 50.86 & 54.06 \\
MemGAS & 49.58 & 52.70 & 53.58 & \uline{49.46} & 50.98 & 72.06 & 33.64 & 47.92 & 47.87 & 58.48 \\
LightMem & 81.52 & 53.19 & \uline{58.12} & 31.25 & 68.32 & 78.76 & 46.11 & \uline{51.17} & 62.41 & 66.01 \\
GAM & 86.02 & 63.12 & 48.75 & 48.50 & \uline{71.72} & 79.52 & 38.62 & 43.03 & 52.20 & 62.72 \\
CoM & 81.28 & \uline{64.54} & 50.00 & 39.58 & 69.10 & 79.81 & 53.27 & 51.04 & 63.12 & \uline{67.91} \\
xMemory & 79.38 & 53.90 & 38.75 & 41.67 & 63.89 & 65.52 & 52.65 & 43.75 & 50.71 & 57.70 \\
SimpleMem & 75.37 & 61.57 & 47.50 & 45.75 & 65.19 & \uline{79.86} & 39.59 & 49.20 & \uline{65.95} & 65.23 \\
\midrule
\textbf{MemSIF} & \textbf{87.51} & \textbf{68.09} & \textbf{76.32} & \textbf{52.08} & \textbf{79.41} 
                 & \textbf{84.56} & \textbf{68.86} & \textbf{54.61} & \textbf{67.51} & \textbf{74.38} \\
\bottomrule
\end{tabular}%
}
\end{table*}

\section{Dataset Statistics}
\label{app:dataset_stats}

Evaluation is conducted on two long-term memory QA benchmarks, LoCoMo
\cite{maharana2024evaluating} and LongMemEval-S
\cite{wu2024longmemeval}.

LoCoMo contains 10 long-term two-party conversations, each spanning
approximately 27 sessions over 184--293 calendar days and averaging 16.6K
tokens. All questions within the same conversation share the interaction
history, requiring memory systems to answer diverse queries against a single
evolving dialogue. We exclude 446 adversarial
questions and retain 1,540 evidence-grounded evaluation samples.

LongMemEval-S contains 500 independent samples, each paired with a distinct
long interaction history averaging over 100K tokens. After excluding 30
abstention questions that do not require evidence-grounded retrieval, 470
evaluation samples remain. Table~\ref{tab:dataset_stats} summarizes the key
statistics and question category distributions.

\begin{table}[t]
\centering
\caption{Dataset statistics and question category distribution.}
\label{tab:dataset_stats}
\resizebox{\columnwidth}{!}{%
\begin{tabular}{lcc}
\toprule
\textbf{Statistic} & \textbf{LoCoMo} & \textbf{LongMemEval-S} \\
\midrule
Conversations / Samples & 10 & 470 \\
Sessions per history & 27.2 (19--32) & 47.7 (38--62) \\
Avg.\ tokens & 16,641 & 103,030 \\
Date span (days) & 238 (184--293) & 28.6 (0--304) \\
\midrule
\multicolumn{3}{l}{\textit{Question Category Distribution}} \\
\midrule
S-hop & 841 (54.6\%) & 150 (31.9\%) \\
M-hop & 282 (18.3\%) & 121 (25.7\%) \\
Temp & 321 (20.8\%) & 127 (27.0\%) \\
Kno & 96 (6.2\%) & 72 (15.3\%) \\
\midrule
\textbf{Total} & \textbf{1,540} & \textbf{470} \\
\bottomrule
\multicolumn{3}{l}{\footnotesize Token counts are computed using the \texttt{cl100k\_base} tokenizer.} \\
\end{tabular}%
}
\end{table}

\paragraph{Question Categories.}
We report all results using four categories, S-hop, M-hop, Temp, and Kno.
S-hop denotes single-hop factual recall, M-hop requires evidence from multiple
segments, Temp involves temporal reasoning, and Kno requires knowledge-based
inference or synthesis. For LongMemEval-S, single-session-user,
single-session-assistant, and single-session-preference are mapped to S-hop;
multi-session to M-hop; temporal-reasoning to Temp; and knowledge-update to
Kno.

\section{Implementation Details}
\label{app:implementation_details}

\subsection{MemSIF Hyperparameters}
\label{app:hyperparameters}

MemSIF computes a matching score for memory units following Equation~(1) in the
main paper:
\begin{equation}
\phi(A,B) = \alpha \cdot s_{\mathrm{sem}}(A,B) + (1-\alpha) \cdot J(\mathcal{K}_A,\mathcal{K}_B),
\end{equation}
where $s_{\mathrm{sem}}$ is cosine similarity normalized to $[0,1]$, and
$J(\mathcal{K}_A,\mathcal{K}_B)$ is the Jaccard overlap between key-entity
sets.

\begin{table}[t]
\centering
\caption{MemSIF hyperparameters (LoCoMo defaults).}
\label{tab:hyperparams}
\resizebox{\columnwidth}{!}{%
\small\setlength{\tabcolsep}{4pt}
\begin{tabular}{lll}
\toprule
\textbf{Component} & \textbf{Parameter} & \textbf{Value} \\
\midrule
Topical Segment & $\alpha$ & 0.8 \\
Topical Segment & $\tau_{\mathrm{merge}}$ & 0.60 \\
Topical Segment & $\tau_{\mathrm{split}}$ & 0.325 \\
Event Trajectory & $\alpha$ & 0.8 \\
Event Trajectory & Top-$K$ & 3 \\
ActiveFact Promotion & $\theta_s$ & 0.45 \\
ActiveFact Promotion & $\theta_q$ & 0.45 \\
\bottomrule
\end{tabular}%
}
\end{table}

Topical Segment construction uses a double-threshold boundary strategy:
scores above $\tau_{\mathrm{merge}}=0.60$ trigger automatic merging, scores
below $\tau_{\mathrm{split}}=0.325$ trigger automatic splitting, and
intermediate scores invoke an LLM judge for resolution. Event Trajectory
aggregation uses the same matching function $\phi$ with $\alpha=0.8$ and
Top-$K=3$, relying on an LLM judge for assignment decisions.

For ActiveFact promotion, a candidate cluster is promoted only when both its
source-support score and query-demand score exceed the thresholds
$\theta_s=\theta_q=0.45$.

In our LoCoMo and LongMemEval-S experiments, the configurable CoreFact schema
uses four fact types, \textit{Identity} (static attributes such
as occupation and location), \textit{Event} (time-anchored occurrences such as
travel and achievements), \textit{Preference} (subjective attitudes across
domains such as food, entertainment, and career), and \textit{Relation} (social
ties such as family, friend, and colleague). The Event category describes fact
content and is distinct from Event Trajectories, which organize conversational
segments. All hyperparameters are
shared across backbones. Hyperparameter values were set through manual tuning
on a held-out validation split, informed by conventions from prior work on
embedding-based memory retrieval systems.

\subsection{Baseline Configurations}
\label{app:baseline_configs}

The eight memory-augmented baselines (Mem0, MemoryOS, LightMem, MemGAS, GAM,
CoM, xMemory, SimpleMem) are run using their official open-source
implementations and recommended hyperparameters, except that embedding-based
components use the unified Qwen3-Embedding-8B encoder for fair comparison. The
two non-memory baselines are configured as follows:

\paragraph{Full-Context.}
The entire interaction history is provided as context, with tail truncation to fit within the backbone model's effective context budget ($\approx$32K tokens after reserving space for the system prompt and answer generation). No memory construction is performed.

\paragraph{Naive RAG.}
Each dialogue turn is treated as an individual retrieval chunk. Chunks are indexed using Qwen3-Embedding-8B with cosine similarity, and the top-100 most relevant chunks are retrieved and concatenated in chronological order, subject to the same context budget as Full-Context.

\paragraph{Model and Retrieval Configuration.}
\label{app:model_config}

Qwen3-series models (32B, 8B, 4B) are deployed locally; DeepSeek-v4-pro is
accessed through its official API. Qwen3-series models and DeepSeek-v4-pro are
run in non-thinking modes, while Llama-3.1-8B-Instruct uses its standard
instruction-following configuration. All backbones share identical
answer-generation parameters (temperature $=0.1$, max tokens $=512$), as
described in Section~\ref{app:evaluation_protocol}.

The unified embedding encoder for all methods is Qwen3-Embedding-8B, which produces 4096-dimensional embeddings and uses cosine similarity as the distance metric.

\paragraph{Software and Hardware Environment.}
All experiments are conducted on Ubuntu 22.04.5 LTS with Python~3.10.0,
PyTorch~2.10.0+cu128, CUDA~12.8, vLLM~0.19.0,
transformers~4.57.0, sentence-transformers~5.1.1, and
FlashAttention~v3 (flashinfer~0.6.6). The hardware comprises
2$\times$ NVIDIA H20 GPUs (96\,GB HBM3 each).

\section{Evaluation Protocol}
\label{app:evaluation_protocol}

All methods share identical answer-generation and GPT-4o judge prompts
(Appendix~\ref{app:prompt_templates}). The only method-specific components are
memory construction and retrieval.

\paragraph{Evaluation Pipeline.}
Evaluation proceeds in three stages per test question. First, each method
constructs and retrieves memory evidence using its own mechanism. Second, the
retrieved evidence is inserted into the answer-generation prompt
provided in Appendix~\ref{app:prompt_templates}, and the backbone LLM produces
an answer. Third, a GPT-4o judge compares the generated answer against the gold
reference using the judge prompt in Appendix~\ref{app:prompt_templates} and
returns a \texttt{CORRECT} or \texttt{WRONG} label.

For LoCoMo, we follow the online query-state protocol used in the main text.
Questions from the same conversation are processed in a fixed randomized order.
Query-local evidence is cleared after each answer, while candidate clusters
and persistent ActiveFact entries are retained within the same conversation and reset
across conversations. Memory updates use only retrieved interaction evidence;
gold answers and judge feedback are never used for memory updates.
LongMemEval-S samples use isolated memory states, as each sample carries an independent interaction history.

\paragraph{Repeated Runs.}
Reported ACC values are averaged over three
independent runs.

\subsection{Validation of the LLM-based Evaluator}
\label{app:judge_validation}

To assess the reliability of GPT-4o as the automatic evaluator, we conduct a
human verification study on a stratified sample of generated answers from the
Qwen3-4B experiments.

\paragraph{Sampling.}
We sample 220 generated answers from each dataset (11 systems $\times$ 20
answers per system), totaling 440 answers, stratified by dataset and method.
Each method contributes the same number of randomly selected answers within
each dataset. Within each stratum, answers are sampled without conditioning on
the GPT-4o decision, thereby retaining the natural distribution of answers
judged as \texttt{CORRECT} and \texttt{WRONG}. The sample covers MemSIF and
all compared methods.

\paragraph{Annotation Protocol.}
Two annotators independently label each generated answer as correct or
incorrect. For each instance, they are shown the question, reference answer,
and generated answer, matching the inputs provided to the GPT-4o judge.
Annotators are blinded to both the method identity and the GPT-4o judgment.
They follow the same evaluation criterion as the automatic judge: an answer is
considered correct only when it directly addresses the question and is
semantically consistent with the essential information in the reference
answer. Disagreements are resolved through discussion to obtain an adjudicated
human label.

\paragraph{Agreement Results.}
Table~\ref{tab:judge_validation} reports inter-annotator agreement and the
agreement between GPT-4o and the adjudicated human labels. We report both
percent agreement and Cohen's $\kappa$ to account for chance agreement.
Human--human agreement establishes an upper bound for the evaluation task:
even expert annotators do not achieve perfect agreement, and GPT-4o--human
agreement should be interpreted relative to this ceiling.

\begin{table}[htbp]
\centering
\small
\setlength{\tabcolsep}{3.5pt}
\caption{Human validation of the GPT-4o evaluator. H--H denotes agreement
between the two human annotators, and GPT--H denotes agreement between GPT-4o
and the adjudicated human labels. Overall aggregates both datasets with equal
weight.}
\label{tab:judge_validation}
\resizebox{\columnwidth}{!}{%
\begin{tabular}{lccccc}
\toprule
\textbf{Dataset} &
\textbf{\# Ans.} &
\textbf{H--H Agr.} &
\textbf{H--H $\kappa$} &
\textbf{GPT--H Agr.} &
\textbf{GPT--H $\kappa$} \\
\midrule
LoCoMo        & 220 & 96.4\% & 0.90 & 93.6\% & 0.82 \\
LongMemEval-S & 220 & 94.5\% & 0.86 & 92.3\% & 0.79 \\
\midrule
Overall       & 440 & 95.5\% & 0.88 & 93.0\% & 0.81 \\
\bottomrule
\end{tabular}%
}
\end{table}

\paragraph{Method-wise and Disagreement Analysis.}
Across all 440 sampled answers, GPT-4o and the adjudicated human labels
disagree on 31 cases (7.0\%). The disagreements are nearly symmetric: 15
cases where GPT-4o accepts answers that humans reject, and 16 cases where
GPT-4o rejects answers that humans accept, indicating no systematic bias
toward either over-acceptance or over-rejection. Across individual methods, GPT-4o--human agreement ranges from
87.5\% to 97.5\%. Agreement for MemSIF is 92.5\%, compared with
87.5\%--97.5\% for the baselines. The most
frequent disagreement patterns involve partially correct answers and
semantically equivalent answers with indirect wording.

\subsection{Statistical Reliability Analysis}
\label{app:bootstrap}

To quantify the uncertainty of the reported accuracy gains, we perform paired
bootstrap resampling between MemSIF and the
strongest baseline in each of the six dataset--backbone settings from the main
results (Tables~\ref{tab:main_results} and~\ref{tab:main_results_lmes}).

For each setting, we first average the per-sample binary correctness judgments
across the three independent runs, yielding a single score in $[0,1]$ for each
method--sample pair. The paired difference for sample $i$ is
$\Delta_i = \mathrm{Acc}_{\textsc{MemSIF}}(i) - \mathrm{Acc}_{\text{baseline}}(i)$,
and the point estimate is the sample mean
$\bar{\Delta} = \frac{1}{n}\sum_{i=1}^{n} \Delta_i$.

We construct the bootstrap distribution of $\bar{\Delta}$ with $B = 10{,}000$
resamples. The resampling strategy respects the dependence structure of each
dataset. For LoCoMo, where multiple questions share the same conversation
history, we use a paired cluster bootstrap: the 10 conversations are resampled
with replacement, and all questions belonging to each sampled conversation are
included. For LongMemEval-S, where each sample carries an independent
interaction history, we use a category-stratified paired bootstrap: within each
of the four question categories (S-hop, M-hop, Temp, Kno), we resample the same
number of questions with replacement. In both strategies, MemSIF and the
baseline share the same resampling indices, preserving the paired structure.
For each bootstrap replicate $b$, we compute $\bar{\Delta}^{(b)}$, and the
95\% confidence interval is obtained via the percentile method as
$[\bar{\Delta}_{(0.025)}, \bar{\Delta}_{(0.975)}]$. Random seeds are set to
$2026 + k$ for the $k$-th comparison.

\begin{table}[t]
\centering
\caption{Paired bootstrap reliability analysis. $\Delta$ACC $=$ ACC$_{\textsc{MemSIF}} -$ ACC$_{\text{baseline}}$. The strongest baseline is selected per dataset--backbone setting. All confidence intervals exclude zero.}
\label{tab:bootstrap}
\resizebox{\columnwidth}{!}{%
\small\setlength{\tabcolsep}{3.5pt}
\begin{tabular}{lllrrr}
\toprule
\textbf{Dataset} & \textbf{Backbone} & \textbf{Baseline} & \textbf{$\Delta$ACC (\%)} & \textbf{95\% CI Low} & \textbf{95\% CI High} \\
\midrule
LoCoMo & Qwen3-4B & CoM & 8.79 & 4.28 & 11.41 \\
LoCoMo & Qwen3-32B & GAM & 5.78 & 1.43 & 7.39 \\
LoCoMo & DeepSeek-v4-pro & SimpleMem & 2.29 & 0.45 & 5.03 \\
LongMemEval-S & Qwen3-4B & CoM & 6.15 & 1.69 & 10.83 \\
LongMemEval-S & Qwen3-32B & CoM & 4.62 & 0.67 & 8.64 \\
LongMemEval-S & DeepSeek-v4-pro & SimpleMem & 2.87 & 0.32 & 6.44 \\
\bottomrule
\end{tabular}%
}
\end{table}

All six confidence intervals exclude zero, confirming that MemSIF's
improvements over the strongest baselines are statistically reliable at the
95\% confidence level. The intervals are wider for LongMemEval-S than for
LoCoMo, reflecting the smaller sample size (470 vs.\ 1,540 questions). The
narrowest intervals occur under DeepSeek-v4-pro on both datasets, consistent
with the smaller point estimates; nevertheless, even these comparisons yield
confidence intervals with positive lower bounds, indicating that the gains are
not attributable to sampling variability alone.

\section{Supplementary Experimental Results}
\label{app:supplementary_results}

To assess whether MemSIF's benefits generalize across model scales and architectures, supplementary experiments are conducted with two additional backbones on LoCoMo, Qwen3-8B, which extends the Qwen3 family to an intermediate scale, and Llama-3.1-8B-Instruct, which represents a different architecture family. Table~\ref{tab:locomo_supplementary} reports the results.

\paragraph{Qwen3-8B Results.}
\label{app:qwen3_8b}

Under Qwen3-8B (Table~\ref{tab:locomo_supplementary}), MemSIF achieves the
highest Total ACC at 79.41\%, surpassing GAM, the best-performing baseline at
71.72\%, by 7.69\%. This result lies between Qwen3-4B at 75.62\% and
Qwen3-32B at 82.99\%, following the increasing trend observed across the
evaluated Qwen3 scales.

The Temp category shows the largest gain, with MemSIF achieving 76.32\% versus GAM's
48.75\%, a 27.57\% advantage. This gap is consistent with the Event Trajectory
design, where linking evidence across sessions provides the
cross-temporal connections that Temp questions require, which session-level memory organization does not offer. MemSIF leads on all four
categories, indicating balanced benefits across question types.

\paragraph{Llama-3.1-8B-Instruct Results.}
\label{app:llama_8b}

Under Llama-3.1-8B-Instruct (Table~\ref{tab:locomo_supplementary}), MemSIF
again achieves the highest Total ACC at 74.38\%, surpassing CoM, the
best-performing baseline at 67.91\%, by 6.47\%. MemSIF leads on all four categories. The M-hop
category shows the largest gain, at 68.86\% versus 57.95\%, while S-hop and
Temp also show consistent gains.

Notably, the Llama backbone exhibits a different category-level profile from
Qwen3, with Kno scores substantially higher than under Qwen3-8B (e.g., MemSIF
achieves 67.51\% vs.\ 52.08\%). This difference reflects
Llama-3.1-8B-Instruct's capacity for knowledge-intensive reasoning. MemSIF
leads all four categories, indicating that the benefits extend across
architectures with varying category-level characteristics.

\paragraph{Cross-Backbone Analysis.}
\label{app:cross_backbone}

Across the 4B, 8B, and 32B Qwen3 backbones on LoCoMo, MemSIF's Total ACC rises
from 75.62\% through 79.41\% to 82.99\%, while maintaining the lead over the
best-performing baseline at every scale. The relative gain over the
best-performing baseline is largest at 4B, at 8.79\% over CoM's 66.83\%, and
narrows to 5.78\% at 32B, suggesting that MemSIF provides stronger marginal
benefits when the backbone's intrinsic reasoning capacity is more limited.

The Llama-3.1-8B result, at 74.38\% with a gain of 6.47\% over the best
baseline, indicates that MemSIF's benefits transfer across architectures,
though the magnitude and category profile vary with the backbone's strengths.

\begin{figure*}[t]
\centering
\includegraphics[width=\textwidth]{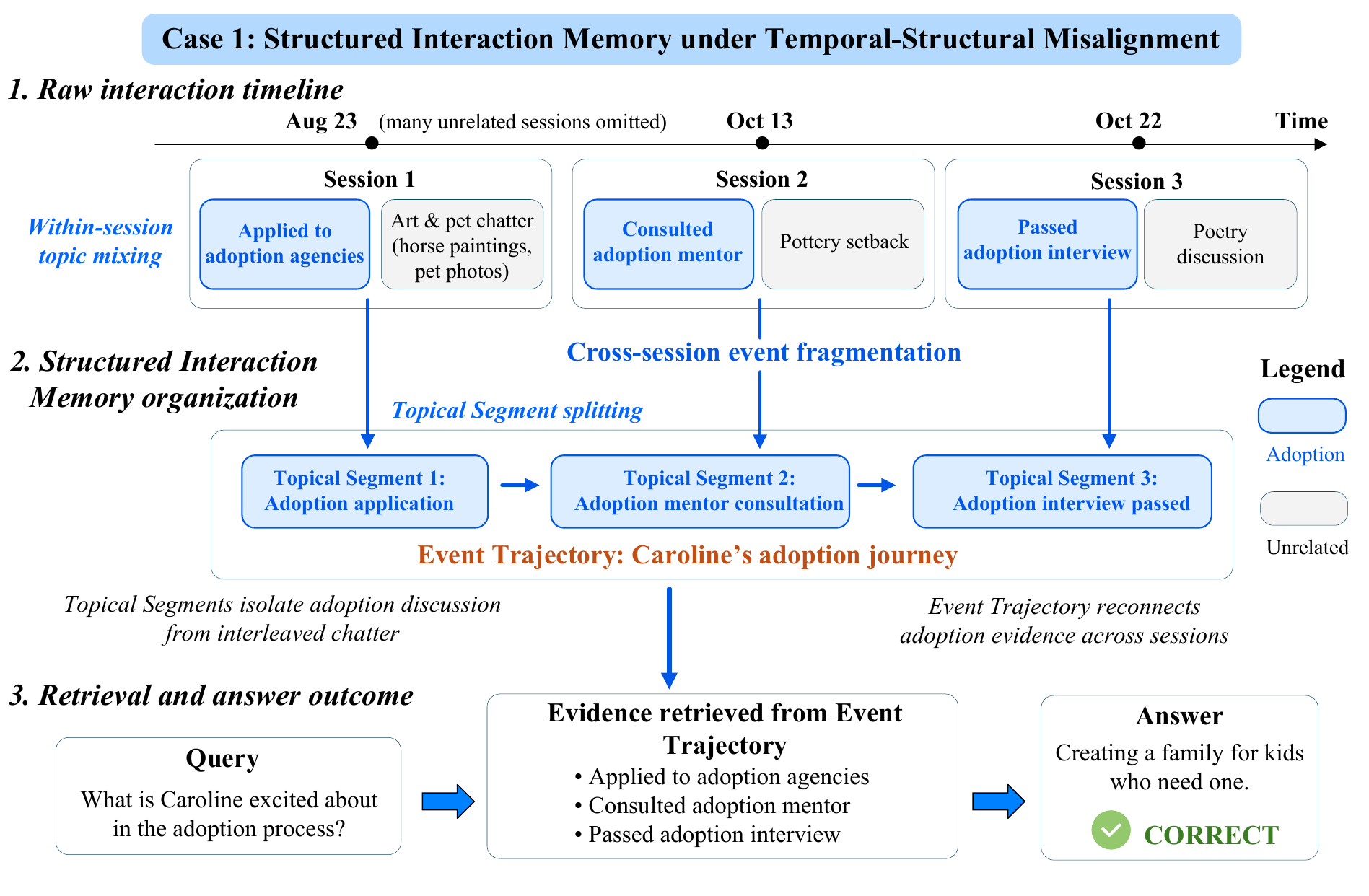}
\caption{Case study 1, Structured Interaction Memory under Temporal--Structural Misalignment (TSM) through Topical Segments and Event Trajectories.}
\label{fig:case_tsm}
\end{figure*}

\begin{figure*}[t]
\centering
\includegraphics[width=\textwidth]{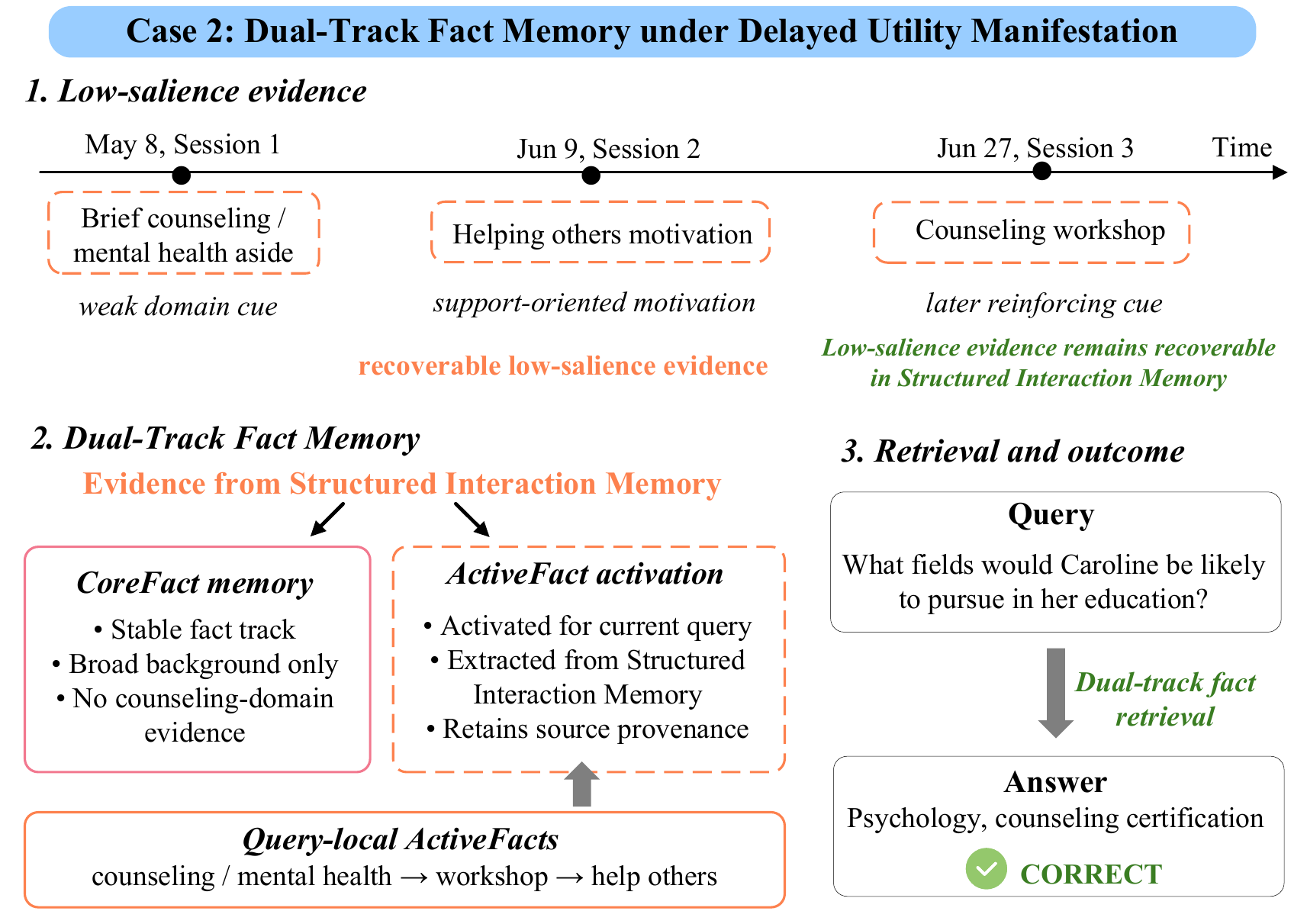}
    \caption{Case study 2, Dual-Track Fact Memory under Delayed Utility Manifestation (DUM) through CoreFact memory and ActiveFact memory.}
\label{fig:case_dum}
\end{figure*}

\section{Sensitivity and Strategy Analysis}
\label{app:sensitivity}

All experiments in this section use LoCoMo with Qwen3-4B and follow the same
retrieval, generation, and judging pipeline as the main experiments
(Appendix~\ref{app:evaluation_protocol}). We focus on construction-time
hyperparameters that affect interaction matching and Topical Segment boundary
decisions under the main LoCoMo online query-state protocol,
the interaction matching weight and the Topical Segment boundary thresholds.
Unless noted otherwise, all other hyperparameters are fixed to their default
values (Appendix~\ref{app:hyperparameters}). Accuracy is reported for four
question categories (S-hop, M-hop, Temp, and Kno), together with the total set.
The default configuration serves as the reference configuration throughout this
section.
\paragraph{Interaction Matching Weight.}

The matching function
$\phi(A,B)=\alpha \cdot s_{\mathrm{sem}}(A,B)+(1-\alpha)\cdot J(\mathcal{K}_A,\mathcal{K}_B)$
combines semantic similarity and entity overlap as complementary signals, where
semantic similarity captures topical coherence, while entity overlap grounds
recurring participants, objects, or events. To evaluate whether this design
is robust to the choice of $\alpha$, a \emph{unified} $\alpha$ is applied to
both Topical Segment construction and Event Trajectory merging, sweeping
$\alpha\in\{0.0,0.25,0.50,0.75,1.0\}$. The Per-Module row uses the actual
per-module defaults, $\alpha=0.8$ for both Topical Segment construction and
Event Trajectory aggregation (Appendix~\ref{app:hyperparameters}).

\begin{table}[t]
\centering
\caption{Sensitivity to the interaction matching weight $\alpha$ on LoCoMo
with Qwen3-4B. ``Per-Module'' uses $\alpha=0.8$ for both Topical Segment
construction and Event Trajectory aggregation.}
\label{tab:sensitivity_alpha}
\resizebox{\columnwidth}{!}{%
\small\setlength{\tabcolsep}{4pt}
\begin{tabular}{l|ccccc}
\toprule
$\alpha$ & S-hop & M-hop & Temp & Kno & Total \\
\midrule
0.00 (entity only) & 78.92 & 75.93 & 62.49 & 49.04 & 73.09 \\
0.25 & 81.37 & 75.95 & 55.20 & 57.71 & 73.45 \\
0.50 & 84.43 & 65.06 & 66.46 & 59.69 & 75.60 \\
0.75 & 85.76 & 65.73 & 48.68 & 60.23 & 72.77 \\
1.00 (semantic only) & 86.25 & 61.11 & 58.20 & 67.82 & 74.65 \\
\midrule
\textbf{Per-Module ($0.8$)}
& \textbf{85.44} & \textbf{61.96} & \textbf{69.49}
& \textbf{50.38} & \textbf{75.62} \\
\bottomrule
\end{tabular}%
}
\end{table}

The results show that semantic similarity and entity overlap are complementary.
The pure-entity setting ($\alpha=0.0$) reaches 73.09\% Total ACC, and the
pure-semantic setting ($\alpha=1.0$) reaches 74.65\%; both are lower than the
balanced unified setting ($\alpha=0.5$, 75.60\%) and the Per-Module default
(75.62\%). Across the unified sweep, Total ACC ranges from 72.77\% to 75.60\%,
indicating that MemSIF is robust to the choice of matching
weight.

Category-level trends further show that the two signals play different roles.
S-hop and Kno improve as $\alpha$ increases, suggesting that single-hop recall
and knowledge-oriented questions benefit more from semantic matching. In
contrast, M-hop peaks under entity-only matching, consistent with cases where
evidence is connected through recurring entities. The Per-Module default gives
the best overall balance across categories.

\paragraph{Segment Boundary Thresholds.}

Topical Segment construction employs a double-threshold strategy, where turns whose
matching score exceeds $\tau_{\mathrm{merge}}$ are auto-merged into the current
segment, turns whose score falls below $\tau_{\mathrm{split}}$ start a new
segment, and cases falling between $\tau_{\mathrm{split}}$ and $\tau_{\mathrm{merge}}$ invoke an LLM judge to
resolve the boundary. This design separates high-confidence merge and split
decisions from ambiguous boundaries, reserving the LLM judge only for
uncertain cases and reducing unnecessary LLM calls. The default thresholds on
LoCoMo are $\tau_{\mathrm{merge}}=0.60$ and $\tau_{\mathrm{split}}=0.325$,
yielding a gap of $0.275$ centered at $c=0.4625$
(Appendix~\ref{app:hyperparameters}).

To assess sensitivity, the gap width
$w=\tau_{\mathrm{merge}}-\tau_{\mathrm{split}}$ is varied while keeping $c$ fixed,
producing five configurations from Very Narrow ($w=0.075$) to Very Wide ($w=0.475$).

\begin{table}[t]
\centering
\caption{Sensitivity to segment boundary thresholds on LoCoMo with Qwen3-4B.
Gap width $w$ is varied while keeping the center $c$ fixed.}
\label{tab:sensitivity_tau}
\resizebox{\columnwidth}{!}{%
\small\setlength{\tabcolsep}{4pt}
\begin{tabular}{lccc|ccccc}
\toprule
Setting & $\tau_{\mathrm{merge}}$ & $\tau_{\mathrm{split}}$ & $w$ & S-hop & M-hop & Temp & Kno & Total \\
\midrule
Very Narrow & 0.50 & 0.425 & 0.075 & 82.23 & 59.18 & 59.01 & 49.55 & 71.13 \\
Narrow      & 0.55 & 0.375 & 0.175 & 86.47 & 62.21 & 54.22 & 52.85 & 73.21 \\
\textbf{Default}     & \textbf{0.60} & \textbf{0.325} & \textbf{0.275} & \textbf{85.44} & \textbf{61.96} & \textbf{69.49} & \textbf{50.38} & \textbf{75.62} \\
Wide        & 0.65 & 0.275 & 0.375 & 79.73 & 65.04 & 59.71 & 54.59 & 71.30 \\
Very Wide   & 0.70 & 0.225 & 0.475 & 80.75 & 64.91 & 63.63 & 53.90 & 72.61 \\
\bottomrule
\end{tabular}%
}
\end{table}

The default setting achieves the highest Total ACC among the tested
configurations, at 75.62\%. Category-level trends reveal trade-offs, where M-hop
improves from Very Narrow to Wide, indicating that allowing a broader ambiguous
region for boundary judgment can help isolate evidence needed for multi-segment
reasoning. Conversely, S-hop drops under wider gaps, indicating a trade-off
between fine-grained evidence separation and preserving local single-hop
continuity. Temp peaks at the default setting, indicating that a moderate gap
best balances these effects for temporal questions. Overall, the default
setting provides the best balance among the tested gaps.

\paragraph{Summary.}
Taken together, these analyses indicate that MemSIF's construction-time choices
remain robust across a range of hyperparameter settings. The matching score
benefits from combining semantic and entity signals, while the double-threshold
segmentation strategy remains effective across reasonable gap-width
variations. These results confirm the complementarity of
the underlying signals.

\begin{table}[t]
\centering
\caption{Amortized efficiency comparison on LoCoMo (Qwen3-4B and Qwen3-32B). \textbf{Bold} indicates the best and \uline{underlined} the second-best per column (higher ACC \,/\, lower tokens and time are better).}
\label{tab:efficiency}
\resizebox{\columnwidth}{!}{%
\small\setlength{\tabcolsep}{4pt}
\begin{tabular}{l|ccc|ccc}
\toprule
\multirow{2}{*}{Method}
  & \multicolumn{3}{c|}{Qwen3-4B}
  & \multicolumn{3}{c}{Qwen3-32B} \\
\cmidrule(lr){2-4} \cmidrule(lr){5-7}
 & ACC (\%) & Tokens & Time (s)
 & ACC (\%) & Tokens & Time (s) \\
\midrule
Full-Context & 62.89 & 11{,}523 & \uline{0.11} & 74.47 & 11{,}525 & \textbf{0.43} \\
Naive RAG & 55.25 & 3{,}532 & \textbf{0.10} & 64.63 & 3{,}533 & \uline{0.64} \\
\midrule
Mem0 & 48.64 & 7{,}056 & 5.90 & 52.47 & 6{,}780 & 17.17 \\
MemoryOS & 50.84 & 2{,}291 & 8.84 & 57.86 & 4{,}822 & 29.29 \\
MemGAS & 56.43 & \uline{1{,}697} & 0.94 & 61.62 & \uline{1{,}650} & 1.33 \\
LightMem & 64.98 & \textbf{1{,}661} & 1.06 & 74.06 & \textbf{1{,}581} & 5.51 \\
GAM & 65.48 & 4{,}064 & 1.30 & \uline{77.21} & 4{,}055 & 8.74 \\
CoM & \uline{66.83} & 2{,}084 & 0.92 & 76.26 & 2{,}078 & 3.65 \\
xMemory & 61.48 & 6{,}897 & 6.70 & 67.70 & 6{,}271 & 17.19 \\
SimpleMem & 61.36 & 2{,}310 & 2.49 & 74.61 & 2{,}293 & 12.31 \\
\midrule
\textbf{MemSIF} & \textbf{75.62} & 3{,}052 & 1.40 & \textbf{82.99} & 2{,}410 & 7.04 \\
\bottomrule
\end{tabular}%
}
\end{table}

\section{Case Studies}
\label{app:case_studies}

Two representative LoCoMo cases illustrate how MemSIF addresses the two
mismatch patterns, where Structured Interaction Memory reconstructs dispersed
evidence under TSM, while Dual-Track Fact Memory extracts low-salience
evidence under DUM.

\paragraph{Case 1: Structured Interaction Memory under TSM.}
\label{app:case_tsm}

Structured Interaction Memory's handling of Temporal--Structural Misalignment (TSM) is examined through the following case. The query \textit{``What is Caroline excited about in the adoption process?''} (gold: \textit{creating a family for kids who need one}) depends on evidence scattered across three sessions over two months, spanning applying to agencies, consulting a mentor, and passing the interview. Within each session, adoption discussion is interleaved with unrelated chatter such as pet photos, horse paintings, and pottery; across sessions, the fragments are separated by weeks of unrelated conversation. This illustrates TSM in its dual form, combining within-session topic mixing with cross-session event fragmentation.

Structured Interaction Memory handles these two patterns through complementary
structures. Topical Segments detect topic boundaries and split conversations
accordingly, isolating adoption discussion from interleaved chatter within the
same session. Event Trajectories then reconnect the separated adoption segments
across sessions, merging them into a single trajectory (\textit{Caroline's
adoption journey}) that recovers the causal chain of apply, consult, and pass.

Figure~\ref{fig:case_tsm} illustrates the reconstructed structure. Once linked
by Event Trajectories, all three adoption fragments are retrieved through their
shared trajectory. This case illustrates why both interaction-level views are
needed. Topical Segments separate adoption-related turns from local topic
mixing, while Event Trajectories reconnect the separated fragments into a
shared retrieval path that spans beyond any fixed chronological window.

\paragraph{Case 2: Dual-Track Fact Memory under DUM.}
\label{app:case_dum}

Dual-Track Fact Memory's handling of Delayed Utility Manifestation (DUM) is
examined through the following case. The query \textit{``What fields would
Caroline be likely to pursue in her education?''} (gold: \textit{Psychology,
counseling certification}) depends on a brief mention of counseling and mental
health that appears in only three conversational turns before the topic shifts
to art. At write time, this aside has low apparent archival value and lies below the typical retention threshold of salience-based
filters. For this query, MemSIF
retrieves the counseling-related evidence from Structured Interaction Memory,
extracts query-local evidence for answering, and can accumulate candidates toward
persistent ActiveFact entries when similar evidence and query demand recur.

CoreFact memory alone cannot answer this query. Its retrieval returns only the generalized preference that Caroline has direction, and cannot identify counseling as the specific domain. CoreFact memory does not encode the aside at write time, since it lacks sufficient salience and cross-session corroboration.

ActiveFact memory fills this gap. When the retrieved
CoreFact entries are insufficient for the query, MemSIF traces back to
Structured Interaction Memory and extracts the counseling-related evidence as
query-local evidence with source provenance. This evidence supports the
current answer and, after answer generation, updates the cross-query
candidate state. When recurring source support and recurring query demand are
observed, the candidate can be promoted to a persistent ActiveFact entry
for later queries.

Figure~\ref{fig:case_dum} illustrates the complementary roles, where CoreFact memory anchors
reusable background knowledge identified at write time, while ActiveFact memory forms facts on demand when
the current question requires them. This case illustrates DUM, where the counseling
aside lies below the salience threshold for CoreFact consolidation, yet
remains recoverable through Structured Interaction Memory and can be extracted
as query-local evidence when the educational-domain query makes it useful.

\section{Efficiency Analysis Details}
\label{app:efficiency_details}

We report amortized cost for each method as the total cost of memory
construction, retrieval, and answer generation divided by the number of test
queries $N_q$. Embedding computation and the external evaluation judge are
excluded. This captures both the one-time memory-building overhead and the
per-query inference cost under the same scope as the main text.
Table~\ref{tab:efficiency} reports token and runtime cost per query alongside
Total ACC for all methods under Qwen3-4B and Qwen3-32B on LoCoMo
($N_q = 1{,}540$).

Under Qwen3-32B, MemSIF achieves the highest ACC at 82.99\% while consuming
2.41K tokens and 7.04 seconds per query. Compared with GAM, the strongest
accuracy baseline at 77.21\%, MemSIF uses 40.6\% fewer tokens (2.41K vs.\
4.06K) and 19.4\% less runtime (7.04s vs.\ 8.74s), while improving ACC by
5.78\%. Compared with SimpleMem at a similar token budget (2.29K vs.\ 2.41K),
MemSIF reduces runtime by 42.8\% and improves ACC by 8.38\%.

LightMem and MemGAS achieve the lowest token consumption under Qwen3-32B
(1.58K--1.65K tokens), but their ACC remains below MemSIF, indicating that aggressive compression trades evidence completeness for
token efficiency.

Full-Context and Naive RAG are fast in wall-clock time because they bypass
memory construction, but both achieve lower ACC than MemSIF.
Full-Context incurs the highest token cost at 11.5K per query, while Naive RAG
has moderate token consumption but lower accuracy than most explicit-memory
methods. These results confirm that explicit memory organization yields higher
accuracy than full-context or simple retrieval baselines, even when those
baselines are computationally efficient.

\section{Prompt Templates}
\label{app:prompt_templates}

This section reports the prompt templates used by MemSIF's LLM-based modules
and by the shared answer-generation and evaluation pipeline. The prompts use
structured outputs where applicable to make intermediate decisions auditable.
Internal MemSIF prompts do not use gold answers, generated answers, or
evaluation feedback.

All methods use the same answer-generation prompt. The context field contains
the evidence retrieved or constructed by each method. Answer generation uses
temperature $0.1$ and max tokens $512$. The GPT-4o judge uses the same
evaluation prompt for all methods with temperature $0.0$ and max tokens $256$.
A generated answer is labeled \texttt{CORRECT} only when it directly answers
the question and is semantically consistent with the essential information in
the gold answer; mere topic or entity overlap is insufficient.

\newcounter{promptlisting}
\renewcommand{\thepromptlisting}{\arabic{promptlisting}}

\newenvironment{prompttemplate}[2][]{%
  \refstepcounter{promptlisting}%
  \begin{tcolorbox}[
    enhanced,
    colback=gray!3,
    colframe=black!55,
    coltitle=white,
    colbacktitle=black!80,
    title={Prompt~\thepromptlisting: #2},
    fonttitle=\bfseries,
    boxrule=0.5pt,
    arc=1mm,
    left=7pt,
    right=7pt,
    top=7pt,
    bottom=7pt,
    boxsep=1pt,
    toptitle=3pt,
    bottomtitle=3pt,
    lefttitle=7pt,
    righttitle=7pt,
    before upper={
      \small
      \linespread{1.1}\selectfont
      \setlength{\parindent}{1pt}
      \setlength{\parskip}{1pt}
      \raggedright
    },
    #1
  ]%
}{%
  \end{tcolorbox}%
}

\newcommand{\promptsection}[1]{%
  \par\vspace{4pt}%
  \noindent\textbf{#1}\par\vspace{2pt}%
}

\newcommand{\promptfield}[1]{%
  \par\vspace{2pt}%
  \noindent\textbf{#1}%
}

\begin{figure*}[t]
\begin{prompttemplate}{Answer Generation Prompt}
\label{lst:answer_prompt}

\promptsection{System Message}
You are a helpful assistant. Answer the question concisely and directly using
only the provided context. If the context contains relevant clues, make a
concise inference grounded only in that context. Do not use external knowledge.
Reply with ``I don't know'' if the context is insufficient.

\promptsection{User Message}
\promptfield{Context:} \texttt{\{\}}

\promptfield{Question:} \texttt{\{\}}

\promptfield{Instruction.}
Provide a direct, short answer based only on the context above. Do not repeat
the question.

\end{prompttemplate}
\end{figure*}

\begin{figure*}[t]
\begin{prompttemplate}{Answer Evaluation Judge Prompt}
\label{lst:judge_prompt}

\promptsection{System Message}
You are an answer evaluation judge. Determine whether a generated answer should
be labeled as \texttt{CORRECT} or \texttt{WRONG}. Return only valid JSON.

\promptsection{User Message}
Evaluate the generated answer against the gold answer for the given question.

\promptfield{Question:} \texttt{\{\}}

\promptfield{Gold Answer:} \texttt{\{\}}

\promptfield{Generated Answer:} \texttt{\{\}}

\promptsection{Evaluation Criteria}
The question asks about information grounded in prior interaction histories.
The gold answer may be concise, and semantic equivalence does not require exact
lexical overlap.

Label the generated answer as \texttt{CORRECT} only if it directly answers the
question and is semantically consistent with the essential information in the
gold answer, including relevant entities, events, relations, attributes,
quantities, or preferences. Mere topic or entity overlap is insufficient.
Additional information is acceptable only if it does not contradict or
materially alter the answer; otherwise, label the answer as \texttt{WRONG}.

For time-related questions, different date formats or relative expressions are
equivalent only when they refer to the same time point or period.

\promptsection{Output Format}
Return only one JSON object with a brief reason and the final label:

\par\vspace{2pt}
{\ttfamily\small
\noindent\{\par
\quad "reason": "...",\par
\quad "label": "CORRECT" or "WRONG"\par
\}\par
}
\vspace{2pt}

Do not include both \texttt{CORRECT} and \texttt{WRONG} in the label field.

\end{prompttemplate}
\end{figure*}

\begin{figure*}[t]
\begin{prompttemplate}{Topical Segment Boundary Judge Prompt}
\label{lst:topical_boundary_prompt}

\promptsection{System Message}
You are a dialogue segmentation judge. Determine whether a new utterance
continues the current Topical Segment or should start a new segment.

\promptsection{User Message}
Given the current segment, a new utterance, and a short look-ahead context,
decide whether the new utterance preserves the same topic or introduces a
topic/event shift.

\promptfield{CURRENT\_SEGMENT:} \texttt{\{current\_segment\}}

\promptfield{NEW\_UTTERANCE:} \texttt{\{new\_utterance\}}

\promptfield{SUBSEQUENT\_TURNS:} \texttt{\{next\_turns\}}

\promptsection{Decision Criteria}
Return \texttt{1} if the new utterance continues the same topic, event, goal,
or discussion direction. Return \texttt{0} if it introduces a different topic,
changes the focus or goal, or if the look-ahead context indicates that the
conversation has shifted. Merge only clear continuations into the current
segment.

\promptsection{Output Format}
Return only \texttt{0} or \texttt{1}.

\end{prompttemplate}
\end{figure*}

\begin{figure*}[t]
\begin{prompttemplate}{Event Trajectory Compatibility Prompt}
\label{lst:event_trajectory_prompt}

\promptsection{System Message}
You are an event-trajectory judge. Determine whether a new Topical Segment
belongs to an existing Event Trajectory.

\promptsection{User Message}
Given an Event Trajectory summary, its existing segment summary, and a new
Topical Segment, decide whether the new segment should be linked to the same
evolving event or task.

\promptfield{EVENT\_TRAJECTORY\_OVERVIEW:} \texttt{\{event\_trajectory\_summary\}}

\promptfield{EXISTING\_SEGMENT\_SUMMARY:} \texttt{\{existing\_segment\_summary\}}

\promptfield{NEW\_TOPICAL\_SEGMENT:} \texttt{\{new\_topical\_segment\_text\}}

\promptsection{Decision Criteria}
Return \texttt{1} if the new segment is a follow-up, progress update,
elaboration, phase transition, decision, or clarification for the same evolving
event or task. Return \texttt{0} if it reflects a different activity, goal,
context, or topic. Link only segments with clear event identity, task
continuity, and state consistency.

\promptsection{Output Format}
Return only \texttt{0} or \texttt{1}.

\end{prompttemplate}
\end{figure*}

\begin{figure*}[t]
\begin{prompttemplate}{CoreFact Construction Prompt}
\label{lst:corefact_construction_prompt}

\promptsection{System Message}
You are a CoreFact construction agent. Extract schema-eligible facts
from Structured Interaction Memory and decide how they should update CoreFact
memory. Return only valid JSON.

\promptsection{User Message}
Given a CoreFact schema, Topical Segment evidence, Event Trajectory evidence,
and existing CoreFacts, construct reusable facts that are supported by the
provided evidence. Topical Segments provide local interaction details, while
Event Trajectories provide cross-time context and corroboration.

\promptfield{COREFACT SCHEMA:} \texttt{\{corefact\_schema\}}

\promptfield{TOPICAL SEGMENT EVIDENCE:} \texttt{\{topical\_segments\}}

\promptfield{EVENT TRAJECTORY EVIDENCE:} \texttt{\{event\_trajectories\}}

\promptfield{EXISTING COREFACTS:} \texttt{\{existing\_corefacts\}}

\promptsection{Rules}
Extract only facts that fit the provided schema and have clear reusable value.
Preserve important names, numbers, titles, dates, relations, and preferences.
Use Event Trajectories to merge or update facts that are supported across
multiple segments. Do not invent, infer, or extend information beyond the
provided evidence. Mark obsolete facts as \texttt{supersede} rather than
deleting them.

\promptsection{Output Format}
Return only one JSON object:

\par\vspace{2pt}
{\ttfamily\small
\noindent\{\par
\quad "corefacts": [\par
\quad\quad \{\par
\quad\quad\quad "operation": "add | update | merge | supersede",\par
\quad\quad\quad "type": "schema type",\par
\quad\quad\quad "subject": "entity or user",\par
\quad\quad\quad "statement": "normalized factual statement",\par
\quad\quad\quad "evidence": ["source ids or short provenance"],\par
\quad\quad\quad "target\_ids": ["existing fact ids, if applicable"]\par
\quad\quad \}\par
\quad ]\par
\}\par
}
\vspace{2pt}

\end{prompttemplate}
\end{figure*}

\begin{figure*}[t]
\begin{prompttemplate}{Evidence Sufficiency Checker Prompt}
\label{lst:sufficiency_checker_prompt}

\promptsection{System Message}
You are an evidence sufficiency assessor. Determine whether the retrieved facts
are sufficient to answer a query. Return only valid JSON.

\promptsection{User Message}
Given a query, its information needs, and the currently retrieved fact summary,
decide whether the summary provides enough evidence to answer the query.

\promptfield{QUERY:} \texttt{\{query\}}

\promptfield{INFORMATION NEEDS:} \texttt{\{needs\_block\}}

\promptfield{RETRIEVED FACT SUMMARY:} \texttt{\{accumulated\_summary\}}

\promptsection{Rules}
Check each information need. Set \texttt{enough=true} only if all needs are
covered clearly and specifically by the retrieved facts. Set
\texttt{enough=false} if any need is missing, vague, off-topic, or unclear. Do
not answer the query, and do not invent or infer facts not explicitly stated in
the retrieved facts.

\promptsection{Output Format}
Return only one JSON object:

\par\vspace{2pt}
{\ttfamily\small
\noindent\{\par
\quad "enough": true or false\par
\}\par
}
\vspace{2pt}

\end{prompttemplate}
\end{figure*}

\begin{figure*}[t]
\begin{prompttemplate}{Query-local Evidence Extraction Prompt}
\label{lst:activefact_extraction_prompt}

\promptsection{System Message}
You are a query-local evidence extraction agent. Select evidence relevant
to a query from the provided sources and distill it into concise query-local evidence.

\promptsection{Stage 1: Evidence Selection Inputs}
\promptfield{RETRIEVED TOPICAL SEGMENTS AND EVENT TRAJECTORIES:} \texttt{\{retrieved\_interaction\_units\}}

\promptfield{SUBJECTS:} \texttt{\{subjects\_block\}}

\promptfield{USER QUERY:} \texttt{\{query\}}

\promptfield{INFORMATION NEEDS:} \texttt{\{needs\_block\}}

\promptsection{Selection Rules}
Select an interaction unit only if its description directly matches the query or
information needs and its \texttt{subject\_id} matches the queried entity. Do
not select a unit merely because it mentions the same person. If no unit is
clearly relevant, return an empty list. Copy selected identifiers verbatim.

\promptsection{Selection Output}
Return only one JSON object:

\par\vspace{2pt}
{\ttfamily\small
\noindent\{\par
\quad "selected": [\par
\quad\quad \{"subject\_id": "copied id", "interaction\_unit\_id": "copied id"\}\par
\quad ]\par
\}\par
}
\vspace{2pt}

\promptsection{Stage 2: Evidence Distillation Inputs}
\promptfield{USER QUERY:} \texttt{\{query\}}

\promptfield{RETRIEVED EVIDENCE:} \texttt{\{facts\_text\}}

\promptsection{Distillation Rules}
Summarize only information grounded in the retrieved evidence and useful for
answering the query. Do not speculate. Keep the summary within five sentences,
omit irrelevant details, use plain prose, and attribute claims to the correct
person when multiple subjects appear. If no evidence helps answer the query,
return exactly
\texttt{NO\_RELEVANT\_EVIDENCE}.

\promptsection{Distillation Output}
Return the summary text only.

\end{prompttemplate}
\end{figure*}

\end{document}